%% file: hippocam.tex
\documentclass{article} %
\usepackage{hippocam,times}
\usepackage[acronym]{glossaries}
\usepackage{xcolor}
\usepackage{graphicx}
\usepackage{booktabs}
\usepackage{colortbl}
\usepackage{array}
\usepackage{tabularx}
\usepackage{placeins}
\usepackage{capt-of}
\usepackage{wrapfig}
\usepackage{needspace}
\usepackage{tikz}
\usepackage{tcolorbox}
\tcbuselibrary{skins,breakable}
\usetikzlibrary{arrows.meta,positioning,calc,fit,backgrounds}

\input{def}

\usepackage[hypertexnames=false]{hyperref}
\usepackage{url}

\usepackage{algorithm}
\usepackage{algpseudocode}
\usepackage{cleveref}
\crefname{figure}{Fig.}{Figs.}
\Crefname{figure}{Figure}{Figures}
\crefname{table}{Table}{Tables}
\Crefname{table}{Table}{Tables}
\crefname{section}{\S}{\S\S}
\Crefname{section}{Section}{Sections}
\crefname{subsection}{\S}{\S\S}
\Crefname{subsection}{Section}{Sections}
\crefname{subsubsection}{\S}{\S\S}
\Crefname{subsubsection}{Section}{Sections}
\crefname{appendix}{\S}{\S\S}
\Crefname{appendix}{Appendix}{Appendices}
\AddToHook{cmd/appendix/after}{%
  \crefalias{section}{appendix}%
  \crefalias{subsection}{appendix}%
  \crefalias{subsubsection}{appendix}%
}
\usepackage{microtype}

\title{Use and Disuse: Intent-Structured Experience Consolidation for Memory and Learning in \glsentryshort{llm} Agents}

\author{%
\textbf{Xiangyi Zeng\thanks{Equal contribution.}\quad Baihang Liu\footnotemark[1]\quad Xutong Wang\quad Ze Jin\quad Yunpeng Li\quad Qixu Liu\thanks{Corresponding author: \texttt{liuqixu@iie.ac.cn}.}}\\[4pt]
{\normalfont\footnotesize Institute of Information Engineering, Chinese Academy of Sciences, Beijing, China}\\
{\normalfont\footnotesize School of Cyber Security, University of Chinese Academy of Sciences, Beijing, China}\\[4pt]
{\normalfont\footnotesize\texttt{\{zengxiangyi,liubaihang,wangxutong,jinze,liyunpeng,liuqixu\}@iie.ac.cn}}
}

\iclrfinalcopy

\begin{document}

\raggedbottom

\maketitle
\lhead{}
\input{abstract}

\input{sections/introduction}
\input{sections/system}

\input{sections/related_work}
\input{sections/experiments}

\input{sections/analysis}
\input{sections/conclusion}

\bibliography{hippocam}
\bibliographystyle{hippocam}

\clearpage
\appendix
\input{sections/experiments/supplement}
\input{sections/experiments/analysis_protocols}
\input{sections/experiments/intent_case}
\input{sections/experiments/analysis_cases}
\input{sections/experiments/stream_cost}
\clearpage
\input{sections/method_appendix}

\end{document}

%% file: def.tex
\newcommand{\sysname}{Hippocam}
\newcommand{\janus}{\textsc{Janus}}
\newcommand{\winnow}{\textsc{Precip}}
\newcommand{\palim}{\textsc{Palim}}
\newcommand{\taskmodel}{DeepSeek-V4.1-Flash}
\newcommand{\judgemodel}{DeepSeek-V4-Pro}
\newacronym{llm}{LLM}{Large Language Model}

\definecolor{expblue}{HTML}{176D8A}
\definecolor{expwarm}{HTML}{9C6545}

\definecolor{expfrozen}{HTML}{DCE8EE}
\definecolor{expsequential}{HTML}{EFE2CC}

%% file: abstract.tex
\begin{abstract}
The evolution of Large Language Model agents from single-task execution to long-term autonomous operation highlights the critical challenge of transforming continuous experiences into reusable knowledge. To address this, we propose \sysname{}, a hierarchical memory and continual learning architecture. \sysname{} draws inspiration from two characteristics of human memory: cognitive processes selectively maintain information relevant to current goals, while long-term memories form gradually through repeated consolidation. Accordingly, \sysname{} structures an agent’s ongoing work as nested intents. The active context remains centered on the current intent, while completed intents are consolidated into the task-relevant outcomes and state needed for subsequent work, rather than carrying forward their full working details. Concurrently, a recursive prefix consolidation mechanism repeatedly consolidates earlier history, causing long-unused experiences to become increasingly abstract. Original interactions are preserved, allowing the agent to progressively recover finer-grained details through the hierarchy and stop once sufficient information is available. Crucially, when past experiences are recalled and reintegrated into active work, they undergo subsequent consolidation alongside new experiences, thereby being reinforced, supplemented, and updated. Through this memory dynamic of use and disuse, \sysname{} connects working context, long-term memory, knowledge accumulation, and skill learning within a single continuously evolving experiential process. This enables agents to learn and evolve capabilities through their own experiences without parameter updates.
\end{abstract}

%% file: sections/introduction.tex
\section{Introduction}
\label{sec:intro}

\gls{llm} agents are increasingly capable of carrying out complex tasks through sustained interaction with their environments \citep{react}.
In long-term, open-ended interactions, enabling agents to continually learn from accumulating experience through memory remains a central challenge for sustained autonomous operation \citep{voyager}.

From the perspective of human cognition, learning through problem solving forms a cycle in which accumulated experience guides the solution of new problems \citep{kolodner1992}.
Memory provides the mechanism through which work and learning become deeply intertwined.
Short-term memory remains tightly focused on the current intent, selectively maintaining the state of the environment and the outcomes and experience acquired so far, thereby enabling sustained focus and progress on the current task \citep{miller2001}.
Long-term memory develops through repeated practice on related tasks and consolidation, as experience accumulates and understanding deepens \citep{ericsson1995,dudai2015}.
When related situations arise, past experience can be reactivated to support understanding, reasoning, and decision-making on new problems \citep{zeithamova2012,shadlen2016}.

\begin{figure}[t]
    \raggedright
    \includegraphics[width=0.85\linewidth]{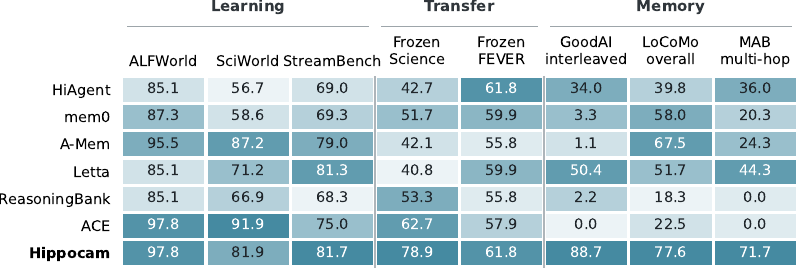}
    \caption{\textbf{Hippocam unifies memory, knowledge use, and learning.} Performance (\%) with \taskmodel{} as the task backbone. Evaluation details appear in \cref{sec:exp}.}
    \label{fig:rq4-profile}
\end{figure}

Inspired by this interplay, we introduce \sysname{}, an \gls{llm} agent that continually learns from its own work through a unified process of organizing and consolidating its interaction history.
\sysname{} does not build a memory bank separate from its interaction history, but remembers and learns by deliberately maintaining its context.
It organizes ongoing work as nested intents, opening an intent when the work develops a more immediate purpose.
As intents close, \sysname{} consolidates their working processes into the outcomes and state relevant to each intent, replacing the original messages in the active context.
As the interaction history grows, \sysname{} recursively consolidates more distant spans, similarly replacing them with the resulting messages.
These consolidated messages can later be consolidated again alongside newly accumulated messages.
This leaves older experience progressively more abstract.
Both forms of consolidation keep the agent's recent context focused and high-signal, enabling accurate and coherent reasoning over long horizons.
They also preserve the original interactions beneath their consolidated representations, naturally organizing the history into a tree.
When subsequent work requires finer evidence, the agent descends selectively through this tree, recovering details until it has enough information to proceed.
\sysname{} draws on past experience to interpret new situations and develop approaches, with the resulting successes and failures providing feedback.
This feedback enters subsequent consolidation together with prior knowledge, allowing \sysname{} to reinforce, supplement, or revise what it has learned and develop broader, reusable insights.
Experience that remains unused continues to recede in detail through recursive consolidation, without separately assigned importance scores or time-based forgetting rules.
This natural dynamic of \emph{use and disuse} unifies the recovery of past experience with learning that deepens what the agent knows, refines how it acts, and expands what it can accomplish.
\sysname{} thereby makes the agent's lifetime of work the engine of continual learning, taking a step toward a unified architecture for memory, knowledge, and learning.

Evaluations show that \sysname{} achieves strong performance across memory, knowledge use, and learning.
Whereas competing methods show more uneven performance across these demands, \sysname{} leads or matches the best compared method on seven of the eight measures in \cref{fig:rq4-profile}.
Ablations and case analyses further show how consolidation and progressive recall support ongoing work and how knowledge develops through use and disuse.
These results demonstrate how our use-and-disuse cycle supports memory and learning through interaction history.

%% file: sections/system.tex
\section{Hippocam}
\label{sec:sys}

\input{sections/system/overview}
\input{sections/system/intent_context}
\input{sections/system/intent_consolidation}
\input{sections/system/prefix_consolidation}
\input{sections/system/progressive_recall}
\input{sections/system/use_disuse}

%% file: sections/system/overview.tex
\subsection{Overview}
\label{sec:overview}

\begin{figure}[t]
    \centering
    \includegraphics[width=0.9\linewidth]{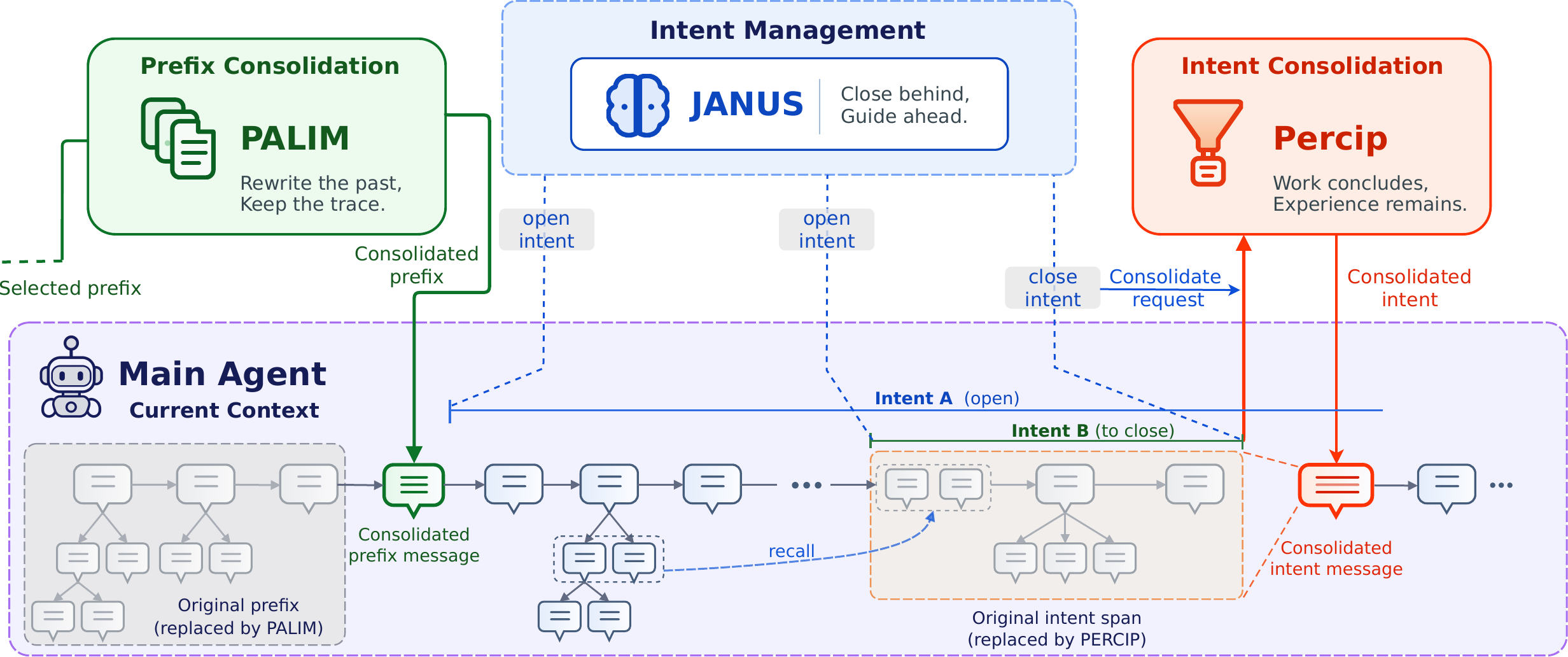}
    \caption{\textbf{Overview of Hippocam.}}
    \label{fig:overview}
\end{figure}

\sysname{} is an \gls{llm} agent that continually learns from its own work by organizing and consolidating its interaction history.
As shown in \cref{fig:overview}, it comprises a main agent responsible for reasoning, tool use, and interaction with the environment, supported by three auxiliary \gls{llm}-based components.
\janus{} maintains the nested intent stack and provides guidance before each main-agent inference (\cref{sec:intent-context}).
\winnow{} performs Intent Consolidation when intents conclude, retaining results and grounded lessons at the level of detail needed for ongoing work (\cref{sec:intent-consolidation}).
\palim{} performs Prefix Consolidation over earlier context, recursively making its content more abstract and less detailed while merging and updating retained knowledge, including acquired facts and learned lessons (\cref{sec:prefix-consolidation}).

Both consolidation mechanisms preserve the messages they replace, forming a message tree from which the main agent can progressively recover finer-grained messages through a recall tool (\cref{sec:progressive-recall}).
Relevant knowledge already in context or recovered through recall can participate in ongoing work and be reinforced, supplemented, or updated through subsequent consolidation, while knowledge not currently used gradually loses detail through recursive Prefix Consolidation.
Together, use and disuse connect task execution, memory, and learning in a continuous cycle (\cref{sec:use-disuse}).

%% file: sections/system/intent_context.tex
\subsection{Intent-Structured Working Context}
\label{sec:intent-context}

\paragraph{Intent stack.}
Let $\mathcal{C}=\langle m_1,\ldots,m_n\rangle$ denote the current interaction context of the main agent.
\sysname{} dynamically organizes $\mathcal{C}$ by \emph{intents}, each of which defines a bounded purpose for the agent's ongoing work.
We represent an intent as $i=\langle p,\chi,b\rangle$, where $p$ states its purpose, $\chi$ specifies its exit conditions, and $b\in\{1,\ldots,n\}$ is the index of the first message governed by the intent.
Recursively nesting more specific sub-intents forms an intent stack $\mathcal{I}=\langle i_1,\ldots,i_d\rangle$, from the most encompassing intent $i_1$ at the bottom to the most immediate $i_d$ at the top.
For each intent $i_j$, all work represented by $m_{b_j},\ldots,m_n$ serves its purpose $p_j$.
The intent stack may begin after the first message, leaving earlier context outside its scope.
Nested intents may begin at the same message, with their order in the stack determining their containment.
This nested chain organizes the multiple purposes jointly served by ongoing work in a form aligned with the linear context of \gls{llm} agents.

\paragraph{Intent maintenance.}
Before each main-agent inference, \janus{} uses the context and intent stack to determine opening and closing operations and generate guidance.
It tracks changes in purpose rather than individual actions, opening nested intents for upcoming work rather than planning the entire task.
Intents close from the top of the stack when contextual evidence supports their success or failure exit conditions.
These operations combine into four forms of intent maintenance.
\emph{Continue} preserves the stack when its intents remain appropriate for the upcoming work.
\emph{Deepen} opens more specific intents within the existing chain.
\emph{Close} removes a concluded suffix without opening new intents.
\emph{Shift} closes a concluded suffix and opens intents for the next direction of work.
\janus{} injects first-person guidance into the main agent's context before its next inference, specifying which purposes have ended, what to pursue next, and the success or failure exit conditions for that work.

%% file: sections/system/intent_consolidation.tex
\subsection{Intent Consolidation}
\label{sec:intent-consolidation}

When \janus{} closes one or more intents, \winnow{} consolidates their work messages into a self-contained first-person message that replaces them in the context.

Given the complete intent stack with closed intents marked, \winnow{} determines what information to retain for subsequent work and at what level of detail.
The resulting message preserves established state changes and conclusions with their concrete supporting evidence and provenance, alongside unprocessed material that enclosing intents may need.
\winnow{} also preserves an impression of the structure and location of consulted material to support targeted inspection in subsequent work.
Informative failures are retained, and facts that jointly support a lesson or pattern are integrated into a shared insight.
The process by which facts were discovered and material that establishes no fact for any intent in the stack are pruned.
This selective consolidation reduces distraction, supports coherent reasoning, and keeps the context focused and efficient, while preserving the information needed to build on established results without repeating prior work.

%% file: sections/system/prefix_consolidation.tex
\subsection{Prefix Consolidation}
\label{sec:prefix-consolidation}

As intents close and new ones open, earlier work messages form a prefix before the new intent stack and become less directly relevant to ongoing work.
The same applies within an intent as work progresses into nested intents, leaving earlier messages at enclosing levels farther from the context's end.
\palim{} consolidates these prefixes to reduce interference from less relevant details and keep the context focused on ongoing work.

\paragraph{Prefix selection.}
The intent stack's opening boundaries partition $\mathcal{C}$ into regions $\mathcal{C}_j=\langle m_{b_j},\ldots,m_{b_{j+1}-1}\rangle$ for $j=0,\ldots,d$, with auxiliary boundaries $b_0=1$ and $b_{d+1}=n+1$.
The region $\mathcal{C}_0$ precedes the stack, while $\mathcal{C}_j$ for $j\geq1$ is local to intent $i_j$.
To preserve details supporting ongoing work, \sysname{} protects a global \emph{working tail}, the shortest suffix containing at least $K$ interaction rounds and $B$ tokens.
Each round comprises a main-agent response and its tool results; consolidated and intent-guidance messages do not count as rounds.
The budget is $B=\min\{\alpha W,\max\{B_{\min},K\operatorname{Quantile}_{q}(\mathcal{R})\}\}$, using the $q$-quantile of interaction-round token counts $\mathcal{R}$, a minimum $B_{\min}$, and a cap $\alpha W$ for context capacity $W$.
After each Intent Consolidation, \palim{} checks the region containing the resulting messages.
It consolidates the portion preceding the working tail only when that portion contains at least $E_{\min}$ intent-consolidated messages and $T_{\min}$ tokens.
Under context pressure, \palim{} consolidates prefixes preceding the working tail in region order $\mathcal{C}_0,\ldots,\mathcal{C}_d$, regardless of accumulation thresholds.
If usage still exceeds capacity, consolidation extends into the working tail.

\paragraph{Recursive consolidation.}
Inspired by the distinction between episodic and semantic memory in human cognition \citep{tulving1972episodic,conway2001sensory}, \palim{} separates each consolidated message into narrative and knowledge.
The narrative recounts the main agent's work and its outcomes, while knowledge comprises acquired facts and lessons extracted by \winnow{} from that work rather than broader lessons generated through consolidation alone.
Expressing knowledge directly about its subjects unbinds it from the original narrative, supporting reasoning and decisions beyond specific circumstances.
Earlier consolidated messages are recursively consolidated with new work, becoming progressively more abstract and less detailed.
Related knowledge is merged, while new facts and lessons supplement or update existing knowledge.

%% file: sections/system/progressive_recall.tex
\subsection{Message Tree and Progressive Recall}
\label{sec:progressive-recall}

Consolidation preserves the original messages and their indices as ordered children of the messages that replace them, forming a message tree whose first layer is the current context.
The main agent has access to a recall tool that returns a message's children given its index.
It first draws on information already in context and may use this tool to recover missing details that could affect its reasoning or next action.
After each recall, it reassesses whether to descend further or stop once the available information is sufficient for ongoing work.

%% file: sections/system/use_disuse.tex
\subsection{Learning through Use and Disuse}
\label{sec:use-disuse}

These mechanisms coordinate use and disuse in a continuous cycle through which \sysname{} learns from its work and applies what it learns to improve subsequent work (\cref{fig:use-disuse}).

\paragraph{Use.}
The main agent treats consolidated messages as a continuation of its own prior work, drawing on retained facts and lessons to guide ongoing work.
It develops approaches by assessing this knowledge against the current task and observations, treating new deductions as hypotheses.
Successes and failures provide feedback on their usefulness in the circumstances encountered.

\Needspace{19\baselineskip}
\begingroup
\setlength{\columnsep}{8pt}
\begin{wrapfigure}[14]{r}{0.425\textwidth}
    \vspace{\dimexpr 8pt-\intextsep\relax}
    \centering
    \setlength{\parskip}{0pt}
    \includegraphics[width=\linewidth]{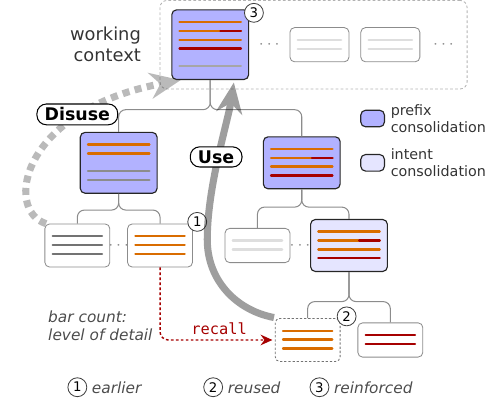}
    \setlength{\abovecaptionskip}{0pt}
    \setlength{\belowcaptionskip}{0pt}
    \caption{How Hippocam learns. Knowledge: \mbox{{\color{orange!85!black}\rule{0.65em}{0.65em}} reused}; \mbox{{\color{red!65!black}\rule{0.65em}{0.65em}} new}; \mbox{{\color{black!55}\rule{0.65em}{0.65em}} fading}.}
    \label{fig:use-disuse}
    \vspace{-5pt}
\end{wrapfigure}

During subsequent Intent Consolidation (\cref{sec:intent-consolidation}), \winnow{} consolidates prior knowledge with these observations and outcomes into grounded facts and lessons.
Later in Prefix Consolidation (\cref{sec:prefix-consolidation}), \palim{} integrates these retained and newly acquired facts and lessons with existing knowledge, reinforcing, supplementing, or updating what the agent has learned.

\paragraph{Disuse.}
In contrast, incidental details and knowledge not currently used are retained in progressively less detail through recursive Prefix Consolidation, reducing their interference with ongoing reasoning and decisions.

The prominence and specificity of retained knowledge therefore evolve through use and disuse, rather than relying on separately assigned importance scores or time-based forgetting rules.
These processes turn the agent's own interaction history into evolving knowledge and capabilities, without maintaining a separate memory bank or skill library.
\par
\endgroup

%% file: sections/related_work.tex
\section{Related Work}
\label{sec:rel}

\paragraph{Context Management.}
Context management is essential for long-horizon agents.
LongLLMLingua~\citep{longllmlingua} compresses context according to query relevance, while SUPO~\citep{supo} learns to condense tool-use histories.
Context-Folding~\citep{context_folding} isolates subtask execution to keep the main context focused.
HiAgent~\citep{hiagent} summarizes completed subgoals, whereas AgentFold~\citep{agentfold} trains the acting model to fold suffixes of its history.
In contrast, \sysname{} not only uses explicit intent requirements rather than inferred future needs to guide suffix consolidation and retention granularity, but goes further by bringing earlier prefixes into context management.

\paragraph{Agent Memory.}
Memory enables agents to maintain continuity and draw on accumulated knowledge across extended interactions.
RAG-based memory organizes information as conversational segments, hierarchies, or associative graphs \citep{secom,raptor,hipporag2}.
Generative Agents~\citep{generative_agents} assigns importance to memories and tracks their recency, while MemoryBank~\citep{memorybank} reinforces recalled memories against time-based forgetting.
Mem0~\citep{mem0} keeps stored facts consistent with new conversations, and A-Mem~\citep{amem} evolves the organization of interconnected memories.
MemTree~\citep{memtree} organizes memories into an evolving tree by semantic similarity.
Other systems track changes over time \citep{theanine,apexmem} or learn to manage memory \citep{mem1,agemem}.
In \sysname{}, context and messages themselves constitute memory, without a separate memory bank or scoring and decay rules.
Unlike MemTree, our tree emerges from the organization and consolidation of ongoing work.

\paragraph{Learning from Experience.}
Recent work increasingly seeks to develop agents that learn continually from experience.
Reflexion~\citep{reflexion} uses verbal feedback to improve subsequent attempts.
Agent Workflow Memory~\citep{awm} accumulates reusable workflows.
ReasoningBank~\citep{reasoningbank} and ACE~\citep{ace} update reasoning memories and playbooks through task feedback.
ETO~\citep{eto} and EMPO\textsuperscript{2}~\citep{empo2} learn from successful and failed trials and memory-guided exploration, respectively, by updating model parameters.
\sysname{} unifies memory and learning through the dynamics of use and disuse, developing knowledge through work and feedback rather than consolidation alone.

%% file: sections/experiments.tex
\section{Experiments}
\label{sec:exp}

We examine \sysname{}'s unified cycle of use and disuse through four research questions.
\textbf{RQ1} Can \sysname{} learn from accumulated experience?
\textbf{RQ2} Can what it learns generalize across task domains?
\textbf{RQ3} Does memory remain reliable over long interactions?
\textbf{RQ4} How general is this unified mechanism?

\paragraph{Experimental setup.}
We compare \sysname{} with mem0, A-Mem, Letta, ReasoningBank, ACE, and HiAgent \citep{mem0,amem,memgpt,reasoningbank,ace,hiagent} across interactive and memory-intensive tasks.
All use \taskmodel{} (1M-token context window) as the task backbone.
We adapt baselines where benchmark support is absent; protocols and implementation details appear in \cref{app:exp-protocols}.

\input{sections/experiments/rq1}
\input{sections/experiments/rq2}
\input{sections/experiments/rq3}

\input{sections/experiments/rq4}

%% file: sections/experiments/rq1.tex
\subsection{RQ1: Learning from Accumulated Experience}
\label{sec:exp-rq1}

We examine whether \sysname{} learns from accumulated experience across three benchmarks covering interactive environments and feedback-driven task streams.
In ALFWorld \citep{alfworld} and ScienceWorld \citep{scienceworld}, agents accumulate experience over an initial task sequence before proceeding to target tasks, retaining experience while the environment resets between episodes (\cref{app:exp-rq1}).
StreamBench \citep{streambench} comprises two task streams, diagnostic classification on DDXPlus and selection among 50 APIs on ToolBench.
Agents receive correctness feedback and a reference answer after each prediction, carrying accumulated experience into subsequent tasks.
We evaluate the first 150 tasks in each stream and average their accuracies equally.

\Needspace{16\baselineskip}
\begin{wraptable}{r}{0.46\textwidth}
\input{tables/experiments/rq1}
\end{wraptable}
\Cref{tab:rq1} reports ALFWorld target-task success, ScienceWorld mean target-task progress, and StreamBench mean accuracy.
Plain uses the same backbone for in-context reasoning without additional memory or learning mechanisms.
Across the three benchmarks, most methods outperform Plain, although some match or fall below it.
\sysname{} improves across all three benchmarks, achieving 97.8\% ALFWorld success and tying ACE for the best result.
Its 81.9\% mean ScienceWorld progress exceeds four of the six compared methods.
On StreamBench, it achieves the highest mean accuracy of 81.7\%, narrowly ahead of Letta's 81.3\%.

\begin{figure}[t]
    \centering
    \includegraphics[width=0.85\linewidth]{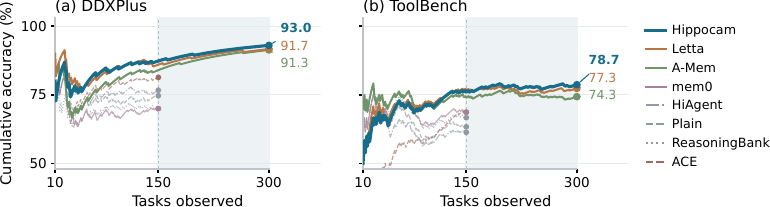}
    \caption{\textbf{Performance as experience accumulates.} Cumulative accuracy on StreamBench. Each stream's top three methods at task 150 continue to task 300. Shading marks this extension.}
    \label{fig:rq1-learning}
    \vspace{-14pt}
\end{figure}

\Cref{fig:rq1-learning} extends each StreamBench stream's three leading methods at task 150 to task 300, tracking cumulative accuracy over all tasks answered so far.
On DDXPlus, \sysname{} continues to improve after early fluctuations and maintains its lead, while Letta and A-Mem also improve.
On ToolBench, \sysname{} makes modest further gains amid fluctuations and moves ahead of Letta, whereas A-Mem shows little sustained improvement in the extended evaluation.
These gains demonstrate more effective learning from accumulated experience than the compared methods.
Detailed scores at both horizons appear in \cref{tab:rq1-stream-detail}.
In both streams, \sysname{} also achieves higher accuracy with about 10\% lower average cost per task in the later 150-task interval (\cref{app:stream-cost}).

%% file: tables/experiments/rq1.tex
\vspace{-\intextsep}\centering
\setlength{\abovecaptionskip}{0pt}
\setlength{\fboxsep}{0.5pt}
\captionof{table}{Task performance (\%) on \mbox{\colorbox{expblue!15}{ALF}World}, \mbox{\colorbox{expblue!15}{Sci}enceWorld}, and \mbox{\colorbox{expblue!15}{Stream}Bench}.}\label{tab:rq1}
\scriptsize\setlength{\tabcolsep}{2.5pt}\renewcommand{\arraystretch}{1.08}
\begin{tabularx}{\linewidth}{>{\hsize=1.6\hsize\linewidth=\hsize\raggedright\arraybackslash}X*{3}{>{\hsize=.8\hsize\linewidth=\hsize\raggedleft\arraybackslash}X}}\toprule
Method & ALF. & Sci. & Stream. \\
\midrule
Plain & 78.4 & 66.9 & 69.0 \\
\midrule
HiAgent & 85.1 & 56.7 & 69.0 \\
mem0 & 87.3 & 58.6 & 69.3 \\
A-Mem & 95.5 & 87.2 & 79.0 \\
Letta & 85.1 & 71.2 & 81.3 \\
ReasoningBank & 85.1 & 66.9 & 68.3 \\
ACE & 97.8 & \textbf{91.9} & 75.0 \\
\midrule
\rowcolor{expblue!7}Hippocam & \textbf{97.8} & 81.9 & \textbf{81.7} \\
\bottomrule\end{tabularx}

%% file: sections/experiments/rq2.tex
\subsection{RQ2: Generalization across Task Domains}
\label{sec:exp-rq2}

We examine whether acquired knowledge generalizes to a new domain and can be integrated with experience accumulated there.
Agents carry memory acquired through household tasks in ALFWorld to scientific experiments in ScienceWorld, and through multi-hop question answering in HotpotQA \citep{hotpotqa} to claim verification in FEVER \citep{fever}.
{\setlength{\fboxsep}{1.5pt}\colorbox{expfrozen}{\emph{Frozen}}} evaluation tests knowledge generalization by evaluating each target-domain task independently from the same source-domain memory snapshot.
{\setlength{\fboxsep}{1.5pt}\colorbox{expsequential}{\emph{Sequential}}} evaluation tests knowledge integration by running the same target-domain task sequence with source-domain memory (\emph{After}) or an initially empty memory (\emph{Fresh}).
In both conditions, experience accumulates across successive target tasks.
The After minus Fresh score difference, $\Delta_{\mathrm{stream}}$, measures whether source-domain knowledge supports subsequent learning in the target domain.
\Cref{tab:rq2} reports mean target-task progress on ScienceWorld and accuracy on FEVER.
Evaluation details appear in \cref{app:exp-rq2}.

\input{tables/experiments/rq2}

\paragraph{Knowledge generalization.}
On ScienceWorld, several methods benefit from source-domain experience, but the gains vary substantially.
\sysname{} achieves 78.9\% mean progress, improving over Plain by 30.1~pp and exceeding the strongest baseline, ACE, by 16.2~pp.
On FEVER, Frozen scores cluster around Plain's 59.3\%, with \sysname{} and HiAgent reaching 61.8\%.
A label-wise analysis suggests limited room for improvement on supported and refuted claims, where Plain already performs strongly, while most errors concern insufficient evidence.
This pattern suggests that source-domain experience provides limited guidance for the remaining target-domain difficulty.

\paragraph{Integrating prior knowledge with new experience.}
Only \sysname{} and ACE achieve positive gains on both domain transitions, while the remaining methods show mixed, unchanged, or reduced performance.
ACE improves ScienceWorld progress by 14.0~pp and FEVER accuracy by 2.0~pp.
\sysname{} achieves larger gains of 19.2 and 6.0~pp, respectively, with a narrower FEVER gap under the scoring sensitivity analysis (\cref{app:exp-rq2}).
These comparisons highlight \sysname{}'s stronger ability to use prior knowledge to support continued learning in a new domain.

%% file: tables/experiments/rq2.tex
\begin{table}[t]
\centering
\setlength{\abovecaptionskip}{0pt}
\caption{\textbf{Cross-domain knowledge generalization and integration.} ScienceWorld mean progress and FEVER accuracy (\%). $\Delta_{\mathrm{stream}}$ reports After minus Fresh in percentage points (pp).}\label{tab:rq2}
\scriptsize
\setlength{\tabcolsep}{3.5pt}
\renewcommand{\arraystretch}{1.13}
\begin{tabular}{lrrl!{\hspace{6pt}}rrl}
\toprule
& \multicolumn{3}{c}{ALFWorld $\rightarrow$ ScienceWorld} & \multicolumn{3}{c}{HotpotQA $\rightarrow$ FEVER} \\
\cmidrule(lr){2-4}\cmidrule(lr){5-7}
Method & \cellcolor{expfrozen}Frozen & \cellcolor{expsequential}Fresh $\to$ After & \cellcolor{expsequential}$\Delta_{\rm stream}$ & \cellcolor{expfrozen}Frozen & \cellcolor{expsequential}Fresh $\to$ After & \cellcolor{expsequential}$\Delta_{\rm stream}$ \\
\midrule
Plain & 48.8 & --- & --- & 59.3 & --- & --- \\
\midrule
HiAgent & 42.7 & 49.6\,$\to$\,49.7 & \setlength{\unitlength}{1mm}\begin{picture}(14,2)\put(7,0){\color{gray!50}\rule{0.15mm}{2mm}}\put(7.000,0.6){\color{expblue}\rule{0.020mm}{0.8mm}}\end{picture}\makebox[7.7mm][r]{\textcolor{expblue}{+0.1}} & \textbf{61.8} & 57.9\,$\to$\,52.1 & \setlength{\unitlength}{1mm}\begin{picture}(14,2)\put(7,0){\color{gray!50}\rule{0.15mm}{2mm}}\put(5.840,0.6){\color{expwarm}\rule{1.160mm}{0.8mm}}\end{picture}\makebox[7.7mm][r]{\textcolor{expwarm}{-5.8}} \\
mem0 & 51.7 & 50.7\,$\to$\,22.6 & \setlength{\unitlength}{1mm}\begin{picture}(14,2)\put(7,0){\color{gray!50}\rule{0.15mm}{2mm}}\put(1.380,0.6){\color{expwarm}\rule{5.620mm}{0.8mm}}\end{picture}\makebox[7.7mm][r]{\textcolor{expwarm}{-28.1}} & 59.9 & 57.9\,$\to$\,57.9 & \setlength{\unitlength}{1mm}\begin{picture}(14,2)\put(7,0){\color{gray!50}\rule{0.15mm}{2mm}}\put(7.000,0.6){\color{gray}\rule{0.000mm}{0.8mm}}\end{picture}\makebox[7.7mm][r]{\textcolor{gray}{+0.0}} \\
A-Mem & 42.1 & 73.9\,$\to$\,65.9 & \setlength{\unitlength}{1mm}\begin{picture}(14,2)\put(7,0){\color{gray!50}\rule{0.15mm}{2mm}}\put(5.400,0.6){\color{expwarm}\rule{1.600mm}{0.8mm}}\end{picture}\makebox[7.7mm][r]{\textcolor{expwarm}{-8.0}} & 55.8 & 55.8\,$\to$\,55.8 & \setlength{\unitlength}{1mm}\begin{picture}(14,2)\put(7,0){\color{gray!50}\rule{0.15mm}{2mm}}\put(7.000,0.6){\color{gray}\rule{0.000mm}{0.8mm}}\end{picture}\makebox[7.7mm][r]{\textcolor{gray}{+0.0}} \\
Letta & 40.8 & 59.3\,$\to$\,54.5 & \setlength{\unitlength}{1mm}\begin{picture}(14,2)\put(7,0){\color{gray!50}\rule{0.15mm}{2mm}}\put(6.040,0.6){\color{expwarm}\rule{0.960mm}{0.8mm}}\end{picture}\makebox[7.7mm][r]{\textcolor{expwarm}{-4.8}} & 59.9 & 69.8\,$\to$\,59.9 & \setlength{\unitlength}{1mm}\begin{picture}(14,2)\put(7,0){\color{gray!50}\rule{0.15mm}{2mm}}\put(5.020,0.6){\color{expwarm}\rule{1.980mm}{0.8mm}}\end{picture}\makebox[7.7mm][r]{\textcolor{expwarm}{-9.9}} \\
ReasoningBank & 53.3 & 48.6\,$\to$\,37.4 & \setlength{\unitlength}{1mm}\begin{picture}(14,2)\put(7,0){\color{gray!50}\rule{0.15mm}{2mm}}\put(4.760,0.6){\color{expwarm}\rule{2.240mm}{0.8mm}}\end{picture}\makebox[7.7mm][r]{\textcolor{expwarm}{-11.2}} & 55.8 & 57.9\,$\to$\,55.8 & \setlength{\unitlength}{1mm}\begin{picture}(14,2)\put(7,0){\color{gray!50}\rule{0.15mm}{2mm}}\put(6.580,0.6){\color{expwarm}\rule{0.420mm}{0.8mm}}\end{picture}\makebox[7.7mm][r]{\textcolor{expwarm}{-2.1}} \\
ACE & 62.7 & 52.5\,$\to$\,66.5 & \setlength{\unitlength}{1mm}\begin{picture}(14,2)\put(7,0){\color{gray!50}\rule{0.15mm}{2mm}}\put(7.000,0.6){\color{expblue}\rule{2.800mm}{0.8mm}}\end{picture}\makebox[7.7mm][r]{\textcolor{expblue}{+14.0}} & 57.9 & 61.8\,$\to$\,63.8 & \setlength{\unitlength}{1mm}\begin{picture}(14,2)\put(7,0){\color{gray!50}\rule{0.15mm}{2mm}}\put(7.000,0.6){\color{expblue}\rule{0.400mm}{0.8mm}}\end{picture}\makebox[7.7mm][r]{\textcolor{expblue}{+2.0}} \\
\rowcolor{expblue!7}Hippocam & \textbf{78.9} & 55.9\,$\to$\,75.1 & \multicolumn{1}{>{\columncolor{expblue!7}[\tabcolsep][\dimexpr\tabcolsep+6pt\relax]}l!{\hspace{6pt}}}{\setlength{\unitlength}{1mm}\begin{picture}(14,2)\put(7,0){\color{gray!50}\rule{0.15mm}{2mm}}\put(7.000,0.6){\color{expblue}\rule{3.840mm}{0.8mm}}\end{picture}\makebox[7.7mm][r]{\textcolor{expblue}{+19.2}}} & \textbf{61.8} & 53.9\,$\to$\,59.9 & \setlength{\unitlength}{1mm}\begin{picture}(14,2)\put(7,0){\color{gray!50}\rule{0.15mm}{2mm}}\put(7.000,0.6){\color{expblue}\rule{1.200mm}{0.8mm}}\end{picture}\makebox[7.7mm][r]{\textcolor{expblue}{+6.0}} \\
\bottomrule
\end{tabular}
\end{table}

%% file: sections/experiments/rq3.tex
\subsection{RQ3: Memory Reliability over Long Interactions}
\label{sec:exp-rq3}

For accumulated experience to support continued learning, agents must remain effective across task switches, reason from past interactions, and retain information as it is updated.
We examine interleaved work in GoodAI LTM \citep{goodai}, reasoning over past conversations in LoCoMo \citep{locomo}, and resolving updated facts in MemoryAgentBench \citep{mab}.
\Cref{app:exp-rq3} provides evaluation protocols and reference configurations.

\begin{figure}[t]
    \centering
    \includegraphics[width=0.85\linewidth]{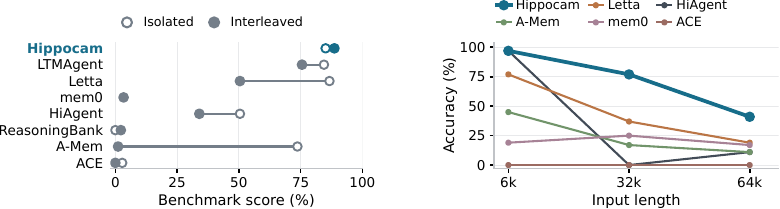}
    \setlength{\abovecaptionskip}{3pt}
    \caption{\textbf{Maintaining usable experience.} \emph{Left:} GoodAI LTM scores under isolated and 32k-span interleaved conditions. \emph{Right:} MemoryAgentBench multi-hop conflict-resolution accuracy across input lengths. Lines connect separate length conditions.}
    \label{fig:rq3-reliability}
\end{figure}

\paragraph{Interleaved work.}
We use GoodAI LTM with a 32k-token memory span for the interleaved condition and compare against isolated evaluation.
We additionally compare against LTMAgent, the long-term memory agent developed by the benchmark authors \citep{goodai}.
LTMAgent, Letta, and HiAgent all lose performance when tasks are interleaved.
In contrast, \sysname{} maintains high performance in both conditions without degradation under interleaving, scoring 85.1\% in isolation and 88.7\% under interleaving (\cref{fig:rq3-reliability}, left).
This contrast highlights \sysname{}'s stronger resilience to interference from interleaved tasks.

\input{tables/experiments/locomo}

\paragraph{Reasoning from past conversations.}
On LoCoMo, \sysname{} achieves 77.6\% overall accuracy, exceeding the strongest baseline, A-Mem, by 10.1~pp (\cref{tab:locomo}).
It leads or ties across all four answerable question categories, with temporal accuracy reaching 86.6\% compared with A-Mem's 67.6\%.
By contrast, ACE's 100\% adversarial accuracy accompanies zero accuracy on answerable questions, reflecting frequent abstention rather than effective recovery of conversational knowledge.
\sysname{} combines stronger conversational reasoning with 87.2\% accuracy on unanswerable questions.

\paragraph{Preserving updated information.}
MemoryAgentBench tests reasoning over revised facts, with conflict-resolution rules provided only at question time.
Across the three input lengths, \sysname{} achieves 71.7\% mean accuracy, compared with 44.3\% for Letta (\cref{fig:rq3-reliability}, right).
It matches HiAgent at 97.0\% on 6k inputs and retains 77.0\% and 41.0\% accuracy at 32k and 64k, where no baseline exceeds 37.0\% and 19.0\%, respectively.
Although longer histories remain challenging, \sysname{} more effectively preserves and applies updated information.

%% file: tables/experiments/locomo.tex
\begin{table}[t]
\centering
\setlength{\abovecaptionskip}{0pt}
\caption{\textbf{What the retained conversation supports.} Binary judge accuracy (\%) on LoCoMo. Evaluation scope is specified in \cref{app:exp-rq3}.}\label{tab:locomo}
\scriptsize
\setlength{\tabcolsep}{3.5pt}
\renewcommand{\arraystretch}{1.06}
\begin{tabular}{lrrrrrr}
\toprule
Method & Overall & Single-hop & Multi-hop & Temporal & Open-domain & Adversarial \\
\midrule
HiAgent & \cellcolor{expblue!10}39.8 & \cellcolor{expblue!9}37.1 & \cellcolor{expblue!7}24.8 & \cellcolor{expblue!6}16.2 & \cellcolor{expblue!11}45.8 & \cellcolor{expblue!15}70.2 \\
mem0 & \cellcolor{expblue!13}58.0 & \cellcolor{expblue!14}67.2 & \cellcolor{expblue!9}37.6 & \cellcolor{expblue!4}8.1 & \cellcolor{expblue!16}77.1 & \cellcolor{expblue!17}85.2 \\
A-Mem & \cellcolor{expblue!14}67.5 & \cellcolor{expblue!16}74.3 & \cellcolor{expblue!9}34.4 & \cellcolor{expblue!14}67.6 & \cellcolor{expblue!13}61.5 & \cellcolor{expblue!16}76.7 \\
Letta & \cellcolor{expblue!12}51.7 & \cellcolor{expblue!11}48.5 & \cellcolor{expblue!7}22.0 & \cellcolor{expblue!9}35.2 & \cellcolor{expblue!11}45.8 & \cellcolor{expblue!18}89.5 \\
ReasoningBank & \cellcolor{expblue!6}18.3 & \cellcolor{expblue!3}2.9 & \cellcolor{expblue!3}0.0 & \cellcolor{expblue!3}2.8 & \cellcolor{expblue!4}7.3 & \cellcolor{expblue!15}72.4 \\
ACE & \cellcolor{expblue!7}22.5 & \cellcolor{expblue!3}0.0 & \cellcolor{expblue!3}0.0 & \cellcolor{expblue!3}0.0 & \cellcolor{expblue!3}0.0 & \cellcolor{expblue!20}100.0 \\
Hippocam & \cellcolor{expblue!16}\textbf{77.6} & \cellcolor{expblue!17}81.5 & \cellcolor{expblue!10}40.8 & \cellcolor{expblue!18}86.6 & \cellcolor{expblue!16}77.1 & \cellcolor{expblue!18}87.2 \\
\bottomrule
\end{tabular}
\end{table}

%% file: sections/experiments/rq4.tex
\subsection{RQ4: Generality of the Unified Memory--Learning Mechanism}
\label{sec:exp-rq4}

\Cref{fig:rq4-profile} reveals distinct capability profiles across the evaluated methods.
Memory-oriented methods support more than recall, with A-Mem performing strongly on ScienceWorld (87.2\%) and Letta on StreamBench (81.3\%).
mem0 also supports learning, transfer, and conversational memory, although it struggles with interleaved work (3.3\%).
By contrast, ACE excels on ScienceWorld (91.9\%) but, like the similarly guidance-oriented ReasoningBank, struggles on LoCoMo (22.5\% and 18.3\%, respectively).
This gap suggests that experience-derived guidance does not necessarily preserve the specific information needed for later recall.
HiAgent exhibits a more balanced profile through subgoal-based context management, suggesting that organizing working context can support a broader range of demands.

\sysname{} combines this breadth with stronger overall performance, leading or tying on seven of the eight measures.
Memory, knowledge development, and learning arise from the same use-and-disuse cycle over its interaction history, demonstrating the broad applicability of a unified mechanism rather than separate capability-specific mechanisms.

%% file: sections/analysis.tex
\section{Analysis}
\label{sec:ana}

\subsection{Ablation Studies}
\label{sec:ablation}

We examine intent-conditioned consolidation, prefix consolidation, and recall in settings that exercise their respective roles.
\Cref{app:analysis-protocols} specifies the interventions and additional comparisons.

\Needspace{21\baselineskip}
\begin{wraptable}{r}{0.39\textwidth}
\input{tables/experiments/ablation}
\end{wraptable}

\paragraph{Intent-conditioned consolidation.}
We remove explicit intent conditioning from consolidation while retaining intent boundaries and Janus guidance.
On GoodAI, filler retention rises from 14\% to 63\%, while scores fall by 23.1~pp under interleaving versus 7.3~pp in isolation (\cref{tab:analysis-ablation}).
A complementary case shows that intent conditioning preserves individual menu options needed for ordering rather than only category ranges (\cref{app:intent-granularity}).
These observations support intent-aware selection of both relevant content and the detail needed for ongoing work.

\paragraph{Prefix consolidation.}
We remove \palim{} and truncate older prefixes when context pressure requires space, while retaining intent consolidation and recall.
ToolBench accuracy falls from 78.7\% to 65.7\% (\cref{tab:analysis-ablation}).
ALFWorld success remains near perfect, but mean actions increase from 9.62 to 11.74, and episode-start context grows from 19.7k to 50.7k tokens in the inspected stream.
Prefix consolidation thus supports more effective subsequent work with a smaller active context.

\paragraph{Recall.}
Disabling recall on the same post-ingestion MemoryAgentBench 64k snapshot reduces accuracy from 41.0\% to 27.0\%.
Access to underlying evidence therefore improves reasoning beyond what the consolidated context alone supports.

\subsection{Learning through Evolving Experience}
\label{sec:analysis-discussion}

\paragraph{Is prior strategy induction necessary?}
The competitive learning performance of memory-oriented methods in \cref{sec:exp-rq4} suggests that retained knowledge can support new work without first being distilled into reusable strategies.
To examine this more directly, we replace Prefix Consolidation with ACE's playbook-generation process in ALFWorld \citep{ace}.
Both configurations achieve near-perfect task success, while the playbook variant uses slightly more actions on average (9.00 versus 8.12).
Inducing a playbook therefore offers no clear advantage here over directly using retained facts and grounded lessons (\cref{tab:analysis-representation}).

\paragraph{Recovering enough detail to proceed.}
In the MemoryAgentBench cases (\cref{fig:analysis-progressive-recall}), one question requires two levels of recall to recover a chain of facts, while another is resolved after one level without opening deeper references.
These cases show that \sysname{} adapts retrieval depth to the task, recovering sufficient evidence without expanding the entire history.

\paragraph{Learning through use and disuse.}
Our analysis of knowledge changes across consecutive consolidations finds that items linked to intervening work are more often supplemented or updated (\cref{app:knowledge-transitions}).
ALFWorld traces show repeated successful desklamp use supporting a lesson across objects and instruction wordings, and later successful searches revising earlier failure accounts to attribute them to search cost (\cref{app:analysis-cases}).
Meanwhile, details of a routine placement episode recede through recursive consolidation (\cref{fig:knowledge-evolution}).
These observations show how use and disuse sustain a continuous cycle of learning through work.

%% file: tables/experiments/ablation.tex
\vspace{-\intextsep}\centering
\setlength{\abovecaptionskip}{0pt}
\captionof{table}{\textbf{Component ablations and consolidation comparison.} Scores, success rates, and accuracy are percentages. Context size is measured in tokens.}\label{tab:analysis-ablation}
\scriptsize
\setlength{\tabcolsep}{2pt}
\renewcommand{\arraystretch}{1.06}
\begin{tabular}{@{}llrr@{}}
\toprule
\textbf{Setting} & \textbf{Metric} & \textbf{\sysname{}} & \textbf{w/o} \\
\midrule
\rowcolor{black!7}\multicolumn{4}{@{}l}{\bfseries Intent conditioning} \\
GoodAI isolated & Score $\uparrow$ & 85.1 & 77.8 \\
\multicolumn{1}{r}{interleaved} & Score $\uparrow$ & 88.7 & 65.6 \\
\midrule
\rowcolor{black!7}\multicolumn{4}{@{}l}{\bfseries Prefix consolidation} \\
ToolBench & Acc. $\uparrow$ & 78.7 & 65.7 \\
ALFWorld & Success $\uparrow$ & 97.8 & 100.0 \\
 & Steps $\downarrow$ & 9.62 & 11.74 \\
 & Context & 19.7k & 50.7k \\
\midrule
\rowcolor{black!7}\multicolumn{4}{@{}l}{\bfseries Recall} \\
MAB-64k & Acc. $\uparrow$ & 41.0 & 27.0 \\
\midrule
\rowcolor{black!7}\multicolumn{3}{@{}l}{\bfseries Prefix consolidation variant} & w/ playbook \\
ALFWorld & Success $\uparrow$ & 100.0 & 97.8 \\
 & Steps $\downarrow$ & 8.12 & 9.00 \\
\bottomrule
\end{tabular}

%% file: sections/conclusion.tex
\section{Conclusion}
\label{sec:con}

We introduced \sysname{}, a unified architecture for memory, knowledge, and learning through the continual consolidation and selective recall of interaction history.
Experiments and analyses demonstrate its broad applicability and illuminate how retained knowledge evolves through use and disuse.
Our findings suggest a path toward agents that continually learn from their own work.

\subsection*{AI use statement}
We used AI tools to accelerate code development, assist literature searches, polish the manuscript, and prepare figures.

%% file: sections/experiments/supplement.tex
\section{Experimental Protocols and Additional Results}
\label{app:exp-protocols}

\paragraph{Scope of the evaluation.}
We report task-level scores for each evaluated configuration, using the same scoring rule across methods within each benchmark.
Results describe single runs.

\paragraph{Backbone and interfaces.}
The task backbone is \taskmodel{}, accessed through the local identifier \texttt{deepseek-flash}. Model-assisted scoring uses \judgemodel{}.
Hippocam uses the same model family for its auxiliary components.
The comparison is not matched by total inference compute or auxiliary output budget.
In interactive environments, observations arrive as tool results and the environment determines completion; reaching an action limit or exhausting the no-action continuation allowance is a failure.
Hippocam's StreamBench configuration exposes its normal tools, whereas the other arms answer through the benchmark interface.
In the QA transfer experiment, Hippocam accesses the same Wikipedia service through a command-line wrapper; the comparison systems use search and lookup calls.
Memory-only Hippocam configurations retain \texttt{recall} while disabling filesystem and shell tools.

\paragraph{Comparison implementations.}
We use the upstream HiAgent code (commit \texttt{cebdd8e}), connected to the task model and environments through an adapter.
The agent is reset only at the beginning of a task stream; its session persists across subsequent tasks and environment episodes.
Its prompt context limit is 8,192 tokens, with output allowances of 4,096 tokens for interactive environments and 16,384 for StreamBench, and a 1,800-second per-task time limit.
The upstream code includes a subsequently added summarization module, so this is an evaluation of that implementation rather than an exact reconstruction of the original paper's code version.

Letta runs with reasoning disabled because of its tool-call protocol.
mem0 and A-Mem disable reasoning during memory extraction; A-Mem also allows up to three attempts at structured extraction.
mem0 uses an expanded extraction output allowance and, in MAB, an adapted general-purpose extraction prompt.
These settings differ from the upstream defaults.

ACE uses the upstream Generator, Reflector, and Curator components (commit \texttt{82709de}) in online mode. Interactive episodes use the environment's action loop followed by reflection and playbook curation. StreamBench uses a 15-task online evaluation window and up to three reflection rounds with the benchmark's feedback. For conversation and fact-ingestion tasks, incoming text passes through generation, reflection, and curation; subsequent questions are answered from the resulting playbook, without an additional fact store. All three roles use \taskmodel{}.

\paragraph{Hippocam consolidation parameters.}
\Cref{tab:hippocam-config} reports the implementation defaults for the consolidation mechanism in \cref{sec:prefix-consolidation}.
The working-tail token budget uses the 95th percentile of observed main-agent action-round lengths, counting each response together with its tool results.
The configured context capacity $W$ includes input and output; pressure control accounts for the current request and reserves space for the next interaction round.
These thresholds control when context is consolidated, rather than assigning importance or decay scores to individual memories.

\begin{table}[htbp]
\centering
\caption{\textbf{Default consolidation configuration.} Token quantities are implementation budgets; $W$ is the configured context capacity.}
\label{tab:hippocam-config}
\footnotesize
\setlength{\tabcolsep}{5pt}
\renewcommand{\arraystretch}{1.08}
\begin{tabular}{lrl}
\toprule
Parameter & Default & Role \\
\midrule
$K$ & 2 & Minimum working-tail interaction rounds \\
$B_{\min}$ & 4,096 tokens & Working-tail token floor \\
$\alpha$ & 0.20 & Working-tail token cap as a fraction of $W$ \\
$q$ & 0.95 & Action-round length quantile \\
$E_{\min}$ & 2 & New intent consolidations in a stable prefix \\
$T_{\min}$ & 16,384 tokens & Stable-prefix size for ordinary consolidation \\
$W$ & 128,000 tokens & Total context capacity \\
Pressure threshold & $0.80W$ & Predicted occupancy that initiates a sweep \\
Capacity threshold & $1.00W$ & Predicted occupancy limit before inference \\
Minimum token reduction & 8,192 tokens & Required saving from a prefix consolidation \\
Compression retries & 2 & Retries when the required reduction is not met \\
\bottomrule
\end{tabular}
\end{table}

\subsection{Experience Accumulation}
\label{app:exp-rq1}

ALFWorld uses distinct source episodes and a 30-action limit per episode.
Target success is measured over unique tasks; preceding memory includes repeated-task experience.

ScienceWorld reports mean target progress, with negative environment scores clipped to zero and percentages normalized to $[0,1]$ before aggregation.
The evaluation pairs source and target variations within the evaluated ScienceWorld task types, with a 50-action limit.
Source and target episodes are kept separate in all reported scores.
\Cref{tab:rq1} averages the two StreamBench accuracies with equal weights over the first 150 tasks in each stream.
Task streams preserve memory across tasks. \Cref{fig:rq1-learning} extends the three highest-scoring methods at task 150 in each stream to task 300. \Cref{tab:rq1-stream-detail} retains all available valid 300-task results, including mem0 and HiAgent on ToolBench. Means use the same horizon for both streams.

\input{tables/experiments/rq1_stream_detail}

\subsection{Source Initialization and Cross-Domain Evaluation}
\label{app:exp-rq2}

The sequential ScienceWorld comparison pairs target identifiers between the source-initialized and empty-memory conditions, with a 30-action limit.
Source memory contains experience from ALFWorld source and target episodes.
HotpotQA$\rightarrow$FEVER transfers experience from source questions to target claims, with a seven-step task limit.
Both conditions accumulate target-domain experience across tasks.
Consequently, their difference is a contrast between two learning streams, not a measure of source-only knowledge on independent tasks.
Sequential differences use paired task identifiers; mem0 and ReasoningBank have narrower matched ScienceWorld coverage than the other methods.

Frozen-source evaluation restores the same source snapshot before each target task and clears the task workspace.
It permits reasoning and memory changes within the current task but prevents those changes from carrying to the next target task.
Plain evaluates target tasks independently without additional mechanisms, providing a reference for Frozen evaluation.
Evaluation costs include target-task calls and within-task memory operations, excluding source experience acquisition.

\paragraph{FEVER scoring sensitivity.}
Under an alternative answer extraction, Hippocam's sequential improvement decreases from six to two percentage points. The main table retains the original scoring.

\input{tables/experiments/rq2_cost}

\subsection{Retention, Interference, and Conflicting Information}
\label{app:exp-rq3}

GoodAI uses the benchmark's own task scores and its isolated and interleaved modes.
Model-assisted scoring steps use \judgemodel{} to extract structured answers for the benchmark's checks. We report the sum of awarded scores divided by the sum of their corresponding maximum scores.
We use the benchmark's 32k memory-span configuration.
The interleaved mode additionally requests extremely brief answers; the modes also differ in reset behavior and wait-action handling.
We interpret the comparison as performance under the two benchmark protocols, without attributing the higher interleaved score to task interleaving itself.

The LoCoMo results cover the full ten-conversation release.
Sessions are presented in order; questions are answered from copies of the post-ingestion state so that answers do not accumulate across questions.
The main table reports binary correctness assigned by \judgemodel{}, not the original benchmark's lexical F1.
Adversarial questions reward identifying unsupported premises, while the remaining categories require substantive answers.
The same judge protocol is used across the reported arms.

For answerable questions, our judge receives the question, reference answer, and extracted model answer.
It accepts semantically equivalent wording, requires dates to match the reference's precision, and requires list answers to contain the same items.
For adversarial questions, it receives the question, the supplied misleading answer, and the model answer; correctness requires identifying that the requested information is absent.
The conversation itself is not included in either judge input.
The judge returns a binary \texttt{CORRECT}/\texttt{WRONG} label after a brief rationale.

MemoryAgentBench uses incremental ingestion followed by independently evaluated questions from the post-ingestion snapshot.
The conflict-resolution results use the full 100-question set at each multi-hop input length.
Answers use the benchmark's substring-matching criterion.
The main Hippocam configuration consolidates at the longer lengths; HiAgent uses its upstream subgoal summaries and prompt-context limit.
These are complete-system comparisons with different retained representations and memory budgets.

DetectiveQA evaluates reasoning over ten novel-length memories containing 71 questions in total; scores count questions rather than ingested documents.

\input{tables/experiments/mab}

The separate 32k-window experiments terminate after capacity failures and are not pooled with the main results.

\paragraph{Mechanistic interpretation.}
The inspected MAB failures include losing updates during consolidation and selecting an obsolete value despite retaining relevant evidence. These observations distinguish information loss from failures to use retained information; the complete-system comparisons do not isolate individual memory mechanisms.

%% file: tables/experiments/rq1_stream_detail.tex
\begin{table}[t]
\centering
\caption{\textbf{StreamBench accuracy at two evaluation horizons.} Mean gives equal weight to the two streams and is reported only when both reach the stated horizon.}\label{tab:rq1-stream-detail}
\footnotesize
\setlength{\tabcolsep}{3.5pt}
\renewcommand{\arraystretch}{1.13}
\begin{tabular}{lrrrrrr}
\toprule
& \multicolumn{3}{c}{First 150 tasks per stream} & \multicolumn{3}{c}{First 300 tasks per stream} \\
\cmidrule(lr){2-4}\cmidrule(lr){5-7}
Method & DDXPlus & ToolBench & Mean & DDXPlus & ToolBench & Mean \\
\midrule
Plain & 74.7 & 63.3 & 69.0 & --- & --- & --- \\
\midrule
HiAgent & 76.7 & 61.3 & 69.0 & --- & 61.3 & --- \\
mem0 & 70.0 & 68.7 & 69.3 & --- & 69.0 & --- \\
A-Mem & 84.0 & 74.0 & 79.0 & 91.3 & 74.3 & 82.8 \\
Letta & 86.0 & \textbf{76.7} & 81.3 & 91.7 & 77.3 & 84.5 \\
ReasoningBank & 70.0 & 66.7 & 68.3 & --- & --- & --- \\
ACE & 81.3 & 68.7 & 75.0 & --- & --- & --- \\
\midrule
\rowcolor{expblue!7}Hippocam & \textbf{87.3} & 76.0 & \textbf{81.7} & \textbf{93.0} & \textbf{78.7} & \textbf{85.8} \\
\bottomrule
\end{tabular}
\par\smallskip
\begin{minipage}{\linewidth}\footnotesize Means weight the two streams equally at the same horizon. Dashes denote unreported results.
\end{minipage}
\end{table}

%% file: tables/experiments/rq2_cost.tex
\begin{table}[t]
\centering
\caption{\textbf{Frozen-source performance and recorded target-evaluation cost.} Performance is reported in percentages; cost is the total USD expenditure for the reported target evaluation, including within-task memory operations and excluding source experience acquisition.}\label{tab:rq2-cost}
\small
\setlength{\tabcolsep}{3.5pt}
\renewcommand{\arraystretch}{1.13}
\begin{tabular}{lrrrr}
\toprule
& \multicolumn{2}{c}{ScienceWorld} & \multicolumn{2}{c}{FEVER} \\
\cmidrule(lr){2-3}\cmidrule(lr){4-5}
Method & Progress & Cost & Accuracy & Cost \\
\midrule
HiAgent & 42.7 & 2.62 & 61.8 & 1.76 \\
mem0 & 51.7 & 0.06 & 59.9 & 0.25 \\
A-Mem & 42.1 & 0.26 & 55.8 & 0.35 \\
Letta & 40.8 & 2.53 & 59.9 & 0.98 \\
ReasoningBank & 53.3 & 0.10 & 55.8 & 0.19 \\
Hippocam & 78.9 & 2.27 & 61.8 & 0.78 \\
\bottomrule
\end{tabular}
\end{table}

%% file: tables/experiments/mab.tex
\begin{table}[t]
\centering
\caption{\textbf{MemoryAgentBench conflict resolution and long-range understanding (LRU).} Accuracy (\%); Overall averages the three multi-hop input lengths.}\label{tab:mab-extra}
\small
\setlength{\tabcolsep}{3.5pt}
\renewcommand{\arraystretch}{1.13}
\begin{tabular}{lrrrrr}
\toprule
Method & 6k & 32k & 64k & Overall & DetectiveQA \\
\midrule
HiAgent & 97.0 & 0.0 & 11.0 & 36.0 & --- \\
Hippocam & 97.0 & 77.0 & 41.0 & 71.7 & 85.9 \\
Letta & 77.0 & 37.0 & 19.0 & 44.3 & 82.4 \\
A-Mem & 45.0 & 17.0 & 11.0 & 24.3 & 78.9 \\
mem0 & 19.0 & 25.0 & 17.0 & 20.3 & --- \\
ACE & 0.0 & 0.0 & 0.0 & 0.0 & --- \\
\bottomrule
\end{tabular}
\par\smallskip
\end{table}

%% file: sections/experiments/analysis_protocols.tex
\subsection{Ablations and Consolidation Comparisons}
\label{app:analysis-protocols}

\paragraph{Intent-conditioning ablation.}
The GoodAI variant removes the intent-specific retention instructions and the explicit intent chain in the consolidation trigger.
Consolidation still occurs at intent boundaries; Janus's focus and completion messages remain in the conversation.
The intervention therefore removes explicit intent conditioning of content selection, while preserving the intent controller and consolidation mechanism.
Both conditions use the same evaluation scope and scoring as the GoodAI comparison in RQ3.
The interleaved score decreases from 88.7\% to 65.6\%, compared with 85.1\% to 77.8\% in isolation.
The paired two-sided sign test for interleaved task scores gives $p=0.11$; each condition is a single run.

We inspect whether filler question--answer pairs remain in the memories that consolidate them; retention increases from 14\% to 63\% without intent conditioning.
This measures retained filler content rather than total memory size.
In SallyAnne cases, the variant incorporates the beginning of a story into a memory associated with another task, while later events remain in the active conversation.
The agent answers with the object's current location rather than the location known to the character; the underlying events remain represented, so this trace alone does not establish that their order has been lost.
Other score differences involve answer behavior or grading, so the aggregate gap cannot be attributed entirely to consolidation.
Excluding the Jokes category, whose grader can credit a list containing the requested joke, preserves the interleaved advantage (91.1\% versus 72.8\%).

\paragraph{Online prefix-consolidation ablation.}
The variant removes prefix consolidation and truncates older prefixes when context pressure requires space, without archiving or consolidating the truncated content.
Intent consolidation and recall remain available.
Each run resumes from the complete system's checkpoint immediately before its first prefix consolidation, using the same task order and driver configuration.
The reported ToolBench score covers the first 300 tasks, including the common initial segment; ALFWorld success and mean actions use the target tasks reported in RQ1.
Context-size statistics describe the first main-agent request of each episode in the inspected ALFWorld stream, including its preceding experience-acquisition segment; they do not measure total inference tokens.

\paragraph{Recall ablation.}
The complete system and the variant answer from copies of the same post-ingestion MemoryAgentBench 64k snapshot.
The variant removes the recall tool and its accompanying instructions; ingestion and stored memory are unchanged.
We report question-paired accuracy and a two-sided exact test on discordant outcomes.
The comparison uses one evaluated answer per question and condition.
Protocol failures following recall remain counted as incorrect in the complete-system score.

\paragraph{Prefix consolidation versus playbook generation.}
The ALFWorld playbook variant replaces Prefix Consolidation with ACE's playbook-generation process, using its reflection and curation components.

Additional comparisons include an empty archive placeholder and a flat summary generated with Letta's summarization prompt.
The latter changes both the representation and its length.

\input{tables/experiments/representation}

Task instances have distinct game files from the preceding experience, although some share a task and room configuration with earlier episodes and vary object placement.
The comparison evaluates within-domain reuse through task success and mean action counts.

%% file: tables/experiments/representation.tex
\begin{table}[t]
\centering
\caption{\textbf{Prefix consolidation and alternative knowledge configurations.} ALFWorld success and mean actions. The ACE variant replaces Prefix Consolidation with its playbook-generation process.}\label{tab:analysis-representation}
\footnotesize
\setlength{\tabcolsep}{7pt}
\renewcommand{\arraystretch}{1.08}
\begin{tabular}{lrr}
\toprule
Configuration & Success (\%) & Mean steps \\
\midrule
Hippocam & 100.0 & 8.12 \\
Empty placeholder & 94.0 & 9.92 \\
ACE playbook & 97.8 & 9.00 \\
Flat summary & 100.0 & 9.68 \\
\bottomrule
\end{tabular}
\end{table}

%% file: sections/experiments/intent_case.tex
\subsection{Intent and the Granularity of Retained Information}
\label{app:intent-granularity}

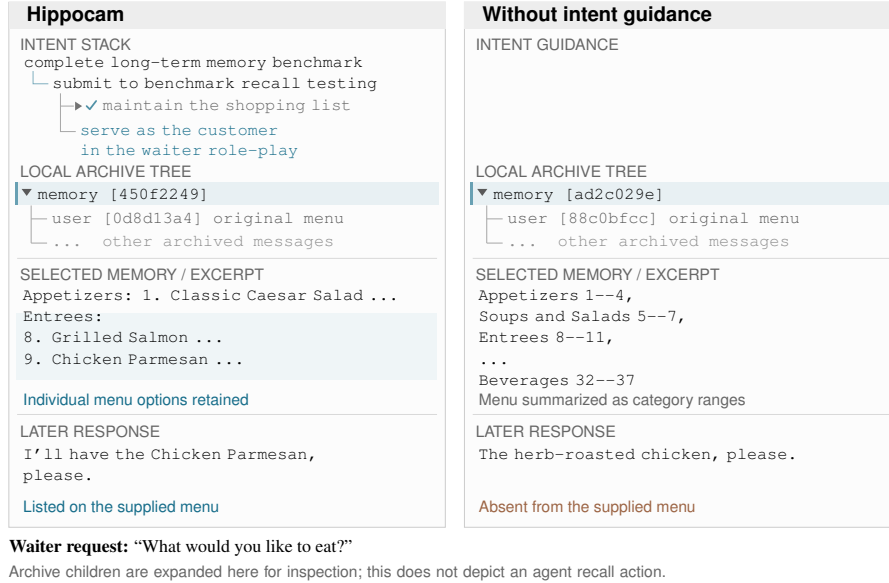
\begin{figure}[htbp]
  \centering
  \resizebox{0.85\linewidth}{!}{\input{figures/experiments/intent_case}}
  \caption{\textbf{Retained detail for a later action.} Local archive subtrees redrawn in the style of the Hippocam viewer, with verbatim memory and response excerpts. The left intent trace shows the completed shopping-list intent and the new customer intent, with labels shortened for display; node IDs identify saved records and ellipses mark omissions. The selected memories retain different levels of menu detail.}
  \label{fig:intent-granularity}
\end{figure}

\Cref{fig:intent-granularity} gives an illustrative Restaurant case outside the aggregate GoodAI comparison.  Both runs received a detailed menu.  Intent-guided consolidation retained the individual dishes, while the variant without intent guidance retained one menu memory chiefly as category names and number ranges.  When later asked to order, the former selected a listed dish and the latter named a dish absent from the menu.  This illustrates a retention-granularity contrast rather than isolating the cause of the response.

%% file: figures/experiments/intent_case.tex
\begin{tikzpicture}[x=1cm,y=1cm,
  pane/.style={draw=black!20,fill=black!1,line width=.4pt},
  mono/.style={font=\ttfamily\fontsize{7.5}{9}\selectfont,anchor=north west,inner sep=0pt},
  labeltext/.style={font=\sffamily\fontsize{7}{8}\selectfont,anchor=north west,inner sep=0pt,text=black!60},
  excerpt/.style={font=\ttfamily\fontsize{7.5}{9.5}\selectfont,anchor=north west,inner sep=0pt,align=left}]
\path[use as bounding box] (0,0) rectangle (14,-9.05);
\foreach \x/\title in {0/Hippocam,7.15/Without intent guidance} {
  \draw[pane] (\x,0) rectangle +(6.85,-8.24);
  \fill[black!7] (\x,0) rectangle +(6.85,-.42);
  \node[anchor=west,font=\sffamily\bfseries\fontsize{9}{10}\selectfont] at (\x+.16,-.21) {\title};
  \node[labeltext] at (\x+.17,-2.56) {LOCAL ARCHIVE TREE};
  \fill[expblue!10] (\x+.10,-2.82) rectangle +(6.65,-.36);
  \fill[expblue!65] (\x+.10,-2.82) rectangle +(.035,-.36);
  \draw[black!35,line width=.45pt] (\x+.35,-3.19) -- (\x+.35,-3.78);
  \draw[black!35,line width=.45pt] (\x+.35,-3.40) -- (\x+.60,-3.40);
  \draw[black!35,line width=.45pt] (\x+.35,-3.78) -- (\x+.60,-3.78);
  \node[mono,text=black!45] at (\x+.67,-3.65) {... other archived messages};
  \draw[black!15] (\x+.13,-4.02) -- (\x+6.72,-4.02);
  \node[labeltext] at (\x+.17,-4.17) {SELECTED MEMORY / EXCERPT};
  \draw[black!15] (\x+.13,-6.47) -- (\x+6.72,-6.47);
  \node[labeltext] at (\x+.17,-6.63) {LATER RESPONSE};
}

\node[labeltext] at (.17,-.56) {INTENT STACK};
\node[mono,text width=6.30cm] at (.23,-.84) {complete long-term memory benchmark};
\draw[expblue!55] (.38,-1.08) -- (.38,-1.28) -- (.62,-1.28);
\node[mono,text width=5.90cm] at (.68,-1.16) {submit to benchmark recall testing};
\draw[black!35] (.81,-1.43) -- (.81,-2.03) -- (1.05,-2.03);
\draw[black!35] (.81,-1.63) -- (1.02,-1.63);
\fill[black!55] (1.04,-1.56) -- (1.04,-1.70) -- (1.15,-1.63) -- cycle;
\draw[expblue!75,line width=.8pt,line cap=round,line join=round]
  (1.22,-1.63) -- (1.27,-1.68) -- (1.37,-1.55);
\node[mono,text=black!60,text width=5.20cm] at (1.47,-1.51) {maintain the shopping list};
\node[mono,text=expblue,text width=5.47cm] at (1.11,-1.91) {serve as the customer\\in the waiter role-play};
\node[labeltext] at (7.32,-.56) {INTENT GUIDANCE};
\foreach \x in {0,7.15} {\fill[black!65] (\x+.20,-2.93) -- (\x+.36,-2.93) -- (\x+.28,-3.04) -- cycle;}
\node[mono] at (.45,-2.91) {  memory [450f2249]};
\node[mono] at (7.60,-2.91) {  memory [ad2c029e]};
\node[mono,text=black!65] at (.67,-3.28) {user [0d8d13a4] original menu};
\node[mono,text=black!65] at (7.82,-3.28) {user [88c0bfcc] original menu};
\fill[expblue!7] (.12,-4.88) rectangle +(6.60,-1.04);
\node[excerpt,text width=6.42cm] at (.22,-4.49) {Appetizers: 1. Classic Caesar Salad \ldots\\
Entrees:\\
8. Grilled Salmon \ldots\\
9. Chicken Parmesan \ldots};
\node[excerpt,text width=6.42cm] at (7.37,-4.49) {Appetizers 1--4,\\
Soups and Salads 5--7,\\
Entrees 8--11,\\
\ldots\\
Beverages 32--37};
\node[labeltext,text=expblue] at (.22,-6.14) {Individual menu options retained};
\node[labeltext] at (7.37,-6.14) {Menu summarized as category ranges};
\node[excerpt,text width=6.40cm] at (.22,-6.98) {I'll have the Chicken Parmesan,\\please.};
\node[excerpt,text width=6.40cm] at (7.37,-6.98) {The herb-roasted chicken, please.};
\node[labeltext,text=expblue] at (.22,-7.82) {Listed on the supplied menu};
\node[labeltext,text=expwarm] at (7.37,-7.82) {Absent from the supplied menu};
\node[anchor=north west,inner sep=0pt,font=\fontsize{8}{10}\selectfont] at (0,-8.43)
  {\textbf{Waiter request:} ``What would you like to eat?''};
\node[labeltext,text width=14cm,align=left] at (0,-8.84)
  {Archive children are expanded here for inspection; this does not depict an agent recall action.};
\end{tikzpicture}

%% file: sections/experiments/analysis_cases.tex
\subsection{Knowledge Changes across Consecutive Consolidations}
\label{app:knowledge-transitions}

We track each knowledge item from one consolidation to the next within a consolidation chain and group these item transitions by evidence of use during the intervening work.
Each transition is classified as unchanged, supplemented, updated, rewritten, or dropped.
Supplementation adds information, updating replaces a concrete value, and rewriting changes wording, trims content, or merges items.
Dropped items are absent from the next consolidated representation, which does not imply deletion of the underlying interactions.
An item can contribute multiple transitions across successive consolidations, so transition counts do not represent distinct items or independent samples.

\paragraph{API-linked use in ToolBench.}
The ToolBench analysis covers two consolidation chains and 3,086 item transitions.
An item is marked as used when an intervening task concerns an API mentioned in the item or the main agent calls that API.
This is an API-linked proxy for use, not a direct observation that the item influenced reasoning.
Used items are divided according to whether the main agent's call exactly matches the reference answer or the reference specifies a different call.
Items mentioning no API remain unclassified rather than being counted as unused.

\input{tables/experiments/knowledge_transitions_toolbench}

Supplementation occurs in 28.4\% of matched-call transitions and 11.7\% of differing-call transitions, compared with 0.9\% of transitions without detected use.
Concrete-value updates are more frequent when the reference specifies a different call (3.4\%) than when the call matches (0.3\%) or no use is detected (0.1\%).
These associations connect subsequent knowledge changes to intervening work and feedback, without establishing that use alone causes the changes.
Most items without detected use remain unchanged (95.8\%), so these adjacent-consolidation statistics do not establish general fading of unused knowledge.

\paragraph{Textual reuse in interactive environments.}
The ALFWorld analysis covers one chain with 23 consolidations, while the ScienceWorld analysis covers five consolidations.
Here the proxy is weaker, marking an item as referenced if a contiguous four-word fragment from it occurs in text written by the main agent between consolidations.
Textual overlap need not imply substantive use, and its absence does not exclude paraphrased or implicit use.
Only unchanged proportions are available for these analyses, so we do not infer how the remaining transitions divide among supplementation, updating, rewriting, and dropping.

\input{tables/experiments/knowledge_transitions_interactive}

In ALFWorld, referenced items remain unchanged less often than items without detected textual reuse (41.6\% versus 63.2\%).
The corresponding ScienceWorld difference is small (57.7\% versus 59.1\%).
The association is therefore clearer in ALFWorld than in ScienceWorld under this textual proxy.

\subsection{Case Evidence for Evolving Memory}
\label{app:analysis-cases}

\begin{figure}[htbp]
    \centering
    \resizebox{0.85\linewidth}{!}{\input{figures/experiments/knowledge_evolution}}
    \caption{\textbf{Experience changes through use and disuse.} Knowledge gains supporting evidence as episode-specific narrative detail recedes. Abridged and paraphrased from ALFWorld traces.}
    \label{fig:knowledge-evolution}
\end{figure}
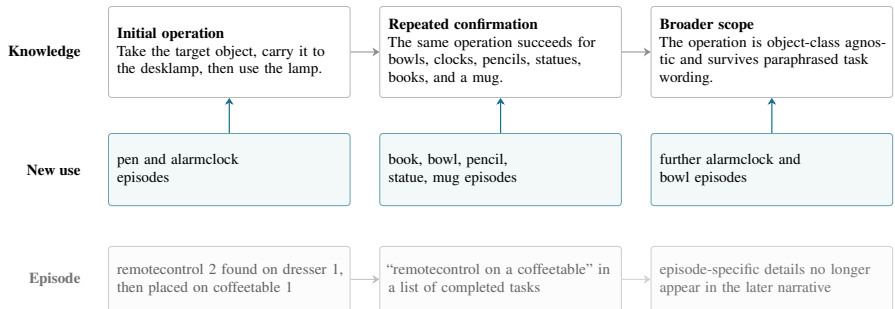

\Cref{fig:knowledge-evolution} follows two ALFWorld memories through repeated prefix consolidation.  The desklamp item begins as a concrete operation, accumulates confirmations from different object classes and phrasings, and is later represented without enumerating those classes.  Meanwhile, an ordinary remotecontrol episode is first retained with its destination, then reduced to an item in a list of completed tasks, after which its episode-specific details no longer occur in the narrative.  Within this run, reusable knowledge persists as the details of individual episodes recede.  The displayed text is abridged and, where quotation marks are absent, paraphrased from the stored memories.

A second lineage shows how new outcomes revise earlier interpretations.  Early searches for an apple and a potato exhausted the action budget without finding either object.  After later episodes found the same object classes and completed related tasks, the retained account came to attribute the earlier failures to search cost.  Further failed and successful searches then entered consolidation alongside this revised account, connecting knowledge updates to continuing experience.

\subsection{Progressive Recall Procedures}

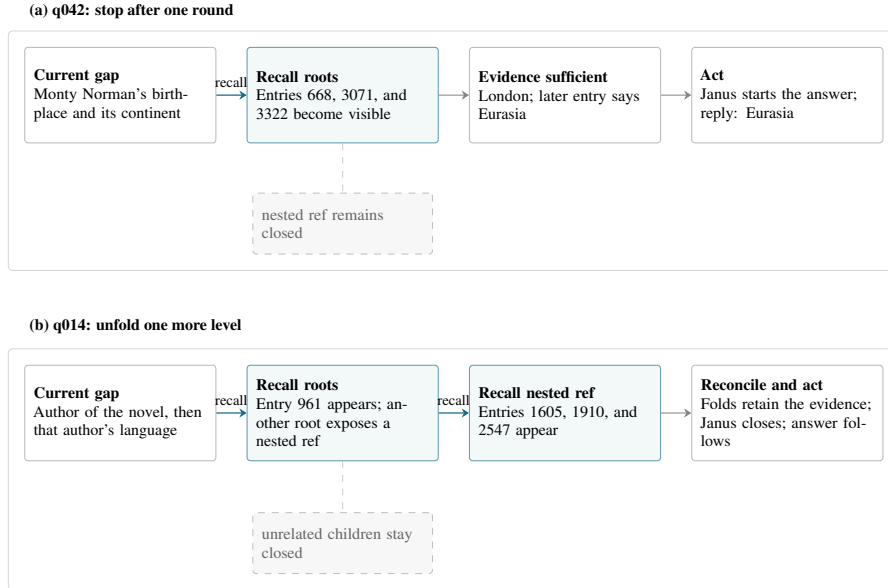
\begin{figure}[htbp]
  \centering
  \resizebox{0.85\linewidth}{!}{\input{figures/experiments/recall_procedure}}
  \caption{Progressive recall in two MemoryAgentBench conflict-resolution cases.  Blue boxes are opened memory; dashed gray boxes are references or children left unopened.  The second case preserves the observed reconsolidation and completion check between recall and answer.}
  \label{fig:analysis-progressive-recall}
\end{figure}

\Cref{fig:analysis-progressive-recall} traces two successful questions from the synthetic facts supplied by MemoryAgentBench; the displayed author, language, birthplace, and continent statements are benchmark records rather than claims about the world.  For q042, one recall round exposes all records needed under the benchmark's later-entry rule, so a deeper reference exposed by the fold remains closed.  For q014, the first round finds one authorship record but leaves the competing authorship and language records behind a nested reference; a second round opens that reference.  Two consolidation folds then carry the recalled evidence into the answering context, and Janus marks the evidence-gathering intentions complete before the response.  Thus the procedure unfolds detail until the current question is supported, while leaving unrelated children unopened.

%% file: tables/experiments/knowledge_transitions_toolbench.tex
\begin{table}[htbp]
\centering
\caption{Knowledge-item transitions in ToolBench grouped by API-linked use between consecutive consolidations. Outcome columns report percentages within each group.}
\label{tab:knowledge-transitions-toolbench}
\footnotesize
\setlength{\tabcolsep}{3pt}
\begin{tabular}{lrrrrr}
\toprule
Intervening use & Unchanged & Supplemented & Updated & Rewritten & Dropped \\
\midrule
No detected use & 95.8 & 0.9 & 0.1 & 3.1 & 0.1 \\
No API, unclassified & 94.1 & 2.1 & 0.3 & 3.4 & 0.1 \\
Used, call matches reference & 57.0 & 28.4 & 0.3 & 14.3 & 0.0 \\
Used, reference call differs & 68.3 & 11.7 & 3.4 & 16.6 & 0.0 \\
\bottomrule
\end{tabular}
\end{table}

%% file: tables/experiments/knowledge_transitions_interactive.tex
\begin{table}[htbp]
\centering
\caption{Knowledge-item transitions grouped by four-word textual reuse in main-agent messages. Unchanged proportions are percentages.}
\label{tab:knowledge-transitions-interactive}
\footnotesize
\setlength{\tabcolsep}{6pt}
\begin{tabular}{llr}
\toprule
Environment & Textual reuse & Unchanged (\%) \\
\midrule
ALFWorld & Detected & 41.6 \\
 & Not detected & 63.2 \\
\midrule
ScienceWorld & Detected & 57.7 \\
 & Not detected & 59.1 \\
\bottomrule
\end{tabular}
\end{table}

%% file: figures/experiments/knowledge_evolution.tex
\begin{tikzpicture}[
  x=1in,y=1in,
  font=\fontsize{7}{8.2}\selectfont,
  >={Stealth[length=3.2pt,width=4.0pt]},
  box/.style={draw=black!28,rounded corners=1.5pt,fill=white,
    align=left,inner sep=4pt,text width=1.43in,minimum height=.58in},
  ev/.style={draw=expblue!65,rounded corners=1.5pt,fill=expblue!5,
    align=left,inner sep=4pt,text width=1.43in,minimum height=.46in},
  arrow/.style={->,draw=black!45,line width=.45pt},
  update/.style={->,draw=expblue,line width=.55pt},
  lane/.style={font=\fontsize{7}{8.2}\selectfont\bfseries,anchor=east}
]
  \node[lane] at (-.08,.51) {Knowledge};
  \node[box] (k1) at (.82,.51) {\textbf{Initial operation}\\Take the target object, carry it to the desklamp, then use the lamp.};
  \node[box] (k2) at (2.55,.51) {\textbf{Repeated confirmation}\\The same operation succeeds for bowls, clocks, pencils, statues, books, and a mug.};
  \node[box] (k3) at (4.28,.51) {\textbf{Broader scope}\\The operation is object-class agnostic and survives paraphrased task wording.};
  \draw[arrow] (k1) -- (k2);
  \draw[arrow] (k2) -- (k3);

  \node[lane] at (-.08,-.24) {New use};
  \node[ev] (e1) at (.82,-.24) {pen and alarmclock\\episodes};
  \node[ev] (e2) at (2.55,-.24) {book, bowl, pencil,\\statue, mug episodes};
  \node[ev] (e3) at (4.28,-.24) {further alarmclock and\\bowl episodes};
  \draw[update] (e1.north) -- (k1.south);
  \draw[update] (e2.north) -- (k2.south);
  \draw[update] (e3.north) -- (k3.south);

  \node[lane,text=black!55] at (-.08,-.94) {Episode};
  \node[box,draw=black!20,text=black!62,fill=black!1,minimum height=.42in] (r1) at (.82,-.94)
    {remotecontrol 2 found on dresser 1, then placed on coffeetable 1};
  \node[box,draw=black!20,text=black!60,fill=black!1,minimum height=.42in] (r2) at (2.55,-.94)
    {``remotecontrol on a coffeetable'' in a list of completed tasks};
  \node[box,draw=black!18,text=black!55,fill=black!1,minimum height=.42in] (r3) at (4.28,-.94)
    {episode-specific details no longer appear in the later narrative};
  \draw[arrow,draw=black!25] (r1) -- (r2);
  \draw[arrow,draw=black!18] (r2) -- (r3);
\end{tikzpicture}

%% file: figures/experiments/recall_procedure.tex
\begin{tikzpicture}[
  x=1in,y=1in,
  font=\fontsize{7}{8.2}\selectfont,
  >={Stealth[length=3.2pt,width=4.0pt]},
  flowbox/.style={draw=black!28,rounded corners=1.5pt,fill=white,align=left,
    inner sep=4pt,text width=1.05in,minimum height=.58in},
  recalled/.style={flowbox,draw=expblue!70,fill=expblue!5},
  closed/.style={flowbox,draw=black!25,fill=black!3,text=black!60,dashed},
  arr/.style={->,draw=black!48,line width=.45pt},
  rec/.style={->,draw=expblue,line width=.55pt},
  panel/.style={draw=black!16,rounded corners=2pt,inner sep=7pt},
  title/.style={font=\fontsize{7}{8.2}\selectfont\bfseries,anchor=west}
]
  \node[title] at (-.05,.76) {(a) q042: stop after one round};
  \node[flowbox] (a0) at (.55,.25) {\textbf{Current gap}\\Monty Norman's birthplace and its continent};
  \node[recalled] (a1) at (1.90,.25) {\textbf{Recall roots}\\Entries 668, 3071, and 3322 become visible};
  \node[flowbox] (a2) at (3.25,.25) {\textbf{Evidence sufficient}\\London; later entry says Eurasia};
  \node[flowbox] (a3) at (4.60,.25) {\textbf{Act}\\Janus starts the answer; reply: Eurasia};
  \draw[rec] (a0) -- node[above,font=\fontsize{6.4}{7.3}\selectfont] {recall} (a1);
  \draw[arr] (a1) -- (a2);
  \draw[arr] (a2) -- (a3);
  \node[closed,text width=.98in,minimum height=.37in] (ac) at (1.90,-.53)
    {nested ref remains closed};
  \draw[draw=black!20,dashed] (a1.south) -- (ac.north);
  \node[panel,fit=(a0)(a1)(a2)(a3)(ac)] {};

  \node[title] at (-.05,-1.15) {(b) q014: unfold one more level};
  \node[flowbox] (b0) at (.55,-1.68) {\textbf{Current gap}\\Author of the novel, then that author's language};
  \node[recalled] (b1) at (1.90,-1.68) {\textbf{Recall roots}\\Entry 961 appears; another root exposes a nested ref};
  \node[recalled] (b2) at (3.25,-1.68) {\textbf{Recall nested ref}\\Entries 1605, 1910, and 2547 appear};
  \node[flowbox] (b3) at (4.60,-1.68) {\textbf{Reconcile and act}\\Folds retain the evidence; Janus closes; answer follows};
  \draw[rec] (b0) -- node[above,font=\fontsize{6.4}{7.3}\selectfont] {recall} (b1);
  \draw[rec] (b1) -- node[above,font=\fontsize{6.4}{7.3}\selectfont] {recall} (b2);
  \draw[arr] (b2) -- (b3);
  \node[closed,text width=.98in,minimum height=.37in] (bc) at (1.90,-2.47)
    {unrelated children stay closed};
  \draw[draw=black!20,dashed] (b1.south) -- (bc.north);
  \node[panel,fit=(b0)(b1)(b2)(b3)(bc)] {};
\end{tikzpicture}

%% file: sections/experiments/stream_cost.tex
\clearpage
\subsection{The Cost of Maintaining and Using Experience}
\label{app:stream-cost}

 Here we examine the cost of maintaining and using that experience on the same StreamBench runs, where both task execution and memory maintenance are metered. We first compare methods over the first 150 tasks, then follow performance and cost through the following tasks.

\paragraph{Accounting.}
We reprice recorded usage with a common schedule: USD 0.00596 per million cached input tokens, 0.29806 per million uncached input tokens, 1.19225 per million output tokens, and 0.07452 per million embedding tokens. Costs include auxiliary model calls and memory writes, and exclude evaluation judges. They are standardized usage costs rather than reconstructed invoices. Input totals sum all calls, including repeated reads of the same history. Letta records total input but not cached input, so we retain its token counts in \cref{tab:stream-cost-extended} without assigning it an exact cost. All other plotted methods use their recorded execution and memory-maintenance usage; their task and auxiliary inference configurations follow the evaluated implementations.

\begin{figure}[htbp]
    \centering
    \includegraphics[width=0.85\linewidth]{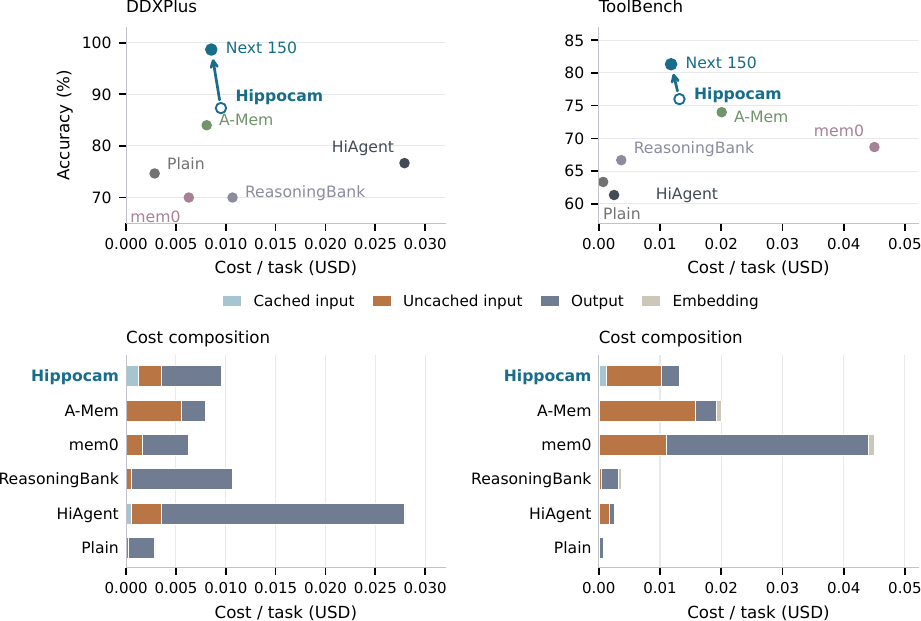}
    \caption{\textbf{Performance, reuse, and the cost of continued work.} \textit{Top:} accuracy versus average cost per task on each stream's first 150 tasks; upper-left is preferable. The blue arrow connects Hippocam's first 150 tasks (open marker) to its next 150 (filled marker); each endpoint reports its own interval. \textit{Bottom:} the first-150 cost decomposed into cached input, uncached input, output, and embeddings. All components include memory maintenance.}
    \label{fig:stream-cost}
\end{figure}

\paragraph{Later work need not cost more per task.}
Across both streams, Hippocam's later interval has higher accuracy and about 10\% lower mean cost per task (\cref{tab:stream-cost-halves}). This occurs despite an increase in total input, showing that repeated use of a growing history need not translate directly into a higher bill. The intervals contain different tasks, so these changes describe the observed runs rather than isolate a causal effect of learning on cost. Across methods, \cref{fig:stream-cost} shows a trade-off: several lower-scoring methods are cheaper, while Hippocam combines higher accuracy with lower cost than A-Mem and mem0 on ToolBench. On DDXPlus, those two methods remain cheaper; the cost advantage is therefore specific to the workload.

\input{tables/experiments/stream_cost_halves}

\paragraph{Stable prefixes make retained experience reusable.}
On ToolBench's first 150 tasks, Hippocam reads 232.5k input tokens per task, compared with A-Mem's 65.0k, but only 30.4k versus 52.7k are uncached. Its cost per correct answer is consequently lower under the stated prices (0.0173 versus 0.0271 USD). Between prefix consolidations, the main agent and Janus append to their respective histories, allowing earlier input to be reused from cache. Intent consolidation also leaves the prefix before the replaced span unchanged. In these implementations, A-Mem and mem0 prepend retrieved memory to the task prompt; changes in that memory can invalidate an otherwise reusable prefix. DDXPlus has shorter, more variable task prompts, leaving less shared input to reuse. These observations connect the cost difference to the organization of context, rather than to its length alone.

Cache reuse does not explain every difference. On ToolBench, approximately 73\% of mem0's cost comes from output tokens, primarily during memory writing. Nor is aggregate cache-hit rate a sufficient efficiency measure: repeatedly reading a long stable history can raise the hit rate without reducing the uncached work. We therefore report absolute token counts and costs per task alongside accuracy. After the next 150 tasks, the cumulative results preserve the main contrast: Hippocam's ToolBench cost per correct answer is 0.0159 USD versus A-Mem's 0.0260, while DDXPlus remains slightly more expensive at 0.0097 versus 0.0091 USD (\cref{tab:stream-cost-extended}).

\input{tables/experiments/stream_cost_extended}

\paragraph{Consolidation has a measurable cost.}
Prefix consolidation rewrites earlier context and interrupts cache reuse. Across both intervals, its own calls account for 34.0\% of Hippocam's ToolBench cost, and the first subsequent main-agent and Janus calls account for another 10.8\%. The corresponding shares on DDXPlus are 26.8\% and 4.2\%. These are costs associated with consolidation, not a counterfactual estimate of its net overhead: the subsequent calls also perform task work. Intent-consolidation calls themselves account for only 0.8\% and 1.3\%, respectively. Thus cache reuse between updates coexists with a substantial cost of periodically reorganizing experience.

\paragraph{Dependence on implementation and pricing.}
These comparisons use the recorded prompt layouts and cache behavior, not optimized implementations of every method. Inference settings also differ: for example, A-Mem's memory-writing calls disable thinking while Hippocam's auxiliary calls enable it. All methods share an API account, so cached prefixes may also be reused across runs. Holding recorded token usage fixed and varying only the cached-input price, Hippocam's lower ToolBench cost per correct answer than A-Mem holds while a cache hit costs less than approximately 15\% of a miss; the reporting schedule uses 2\%. The observed savings therefore depend on substantial cache discounts and do not establish a general cost advantage across providers or tasks.

%% file: tables/experiments/stream_cost_halves.tex
\begin{table}[htbp]
\centering
\caption{\textbf{Later work improves without a higher mean cost per task.} Hippocam's consecutive StreamBench intervals. Input counts sum all agent calls per task, in thousands of tokens; they are not single-request context lengths. Cost per correct answer divides total cost, including incorrect answers, by the number correct.}
\label{tab:stream-cost-halves}
\footnotesize\setlength{\tabcolsep}{4pt}
\begin{tabular}{llrrrrr}
\toprule
Stream & Tasks & Acc. (\%) & Input (k) & Uncached (k) & USD/task & USD/correct \\
\midrule
DDXPlus & First 150 & 87.3 & 207.8 & 8.0 & 0.0095 & 0.0109 \\
 & Next 150 & 98.7 & 252.3 & 8.3 & 0.0085 & 0.0087 \\
\midrule
ToolBench & First 150 & 76.0 & 232.5 & 30.4 & 0.0132 & 0.0173 \\
 & Next 150 & 81.3 & 259.5 & 26.8 & 0.0118 & 0.0145 \\
\bottomrule\end{tabular}
\end{table}

%% file: tables/experiments/stream_cost_extended.tex
\begin{table}[htbp]
\centering
\caption{\textbf{Cumulative cost--performance after the next 150 tasks.} Scores and costs cover both intervals and include memory maintenance. Letta's total input is recorded, but its uncached input and exact repriced cost cannot be recovered.}
\label{tab:stream-cost-extended}
\footnotesize\setlength{\tabcolsep}{4pt}
\begin{tabular}{llrrrrr}
\toprule
Stream & Method & Acc. (\%) & Input (k) & Uncached (k) & USD/task & USD/correct \\
\midrule
DDXPlus & Hippocam & 93.0 & 230.1 & 8.1 & 0.0090 & 0.0097 \\
 & A-Mem & 91.3 & 20.1 & 19.1 & 0.0083 & 0.0091 \\
 & Letta & 91.7 & 77.1 & --- & --- & --- \\
\midrule
ToolBench & Hippocam & 78.7 & 246.0 & 28.6 & 0.0125 & 0.0159 \\
 & A-Mem & 74.3 & 61.9 & 51.6 & 0.0193 & 0.0260 \\
 & mem0 & 69.0 & 48.3 & 38.0 & 0.0470 & 0.0681 \\
 & HiAgent & 61.3 & 7.7 & 5.5 & 0.0024 & 0.0038 \\
 & Letta & 77.3 & 61.3 & --- & --- & --- \\
\bottomrule\end{tabular}
\end{table}

%% file: sections/method_appendix.tex
\section{Core Algorithms and Prompt Specifications}
\label{app:method}

We specify how the components in \cref{sec:overview} interact, followed by selected instructions governing their behavior. The algorithms describe context and archive operations; the prompts determine intent boundaries, retained content, and the agent's use of experience. Default thresholds appear in \cref{tab:hippocam-config}.

\subsection{Execution and Memory Operations}
\label{app:method-algorithms}

The state consists of the active context $\mathcal{C}$, the nested intent stack $\mathcal{I}$, and the archive $\mathcal{A}$. Each consolidation replaces a contiguous span by a new message bearing an archive reference. The replaced messages and their existing subtrees become its ordered children. Intent boundaries following the replaced span are shifted accordingly. This preserves access to the underlying records without requiring their details to remain in the active context.

\begin{algorithm}[H]
\color{black}
\caption{An interaction step and its memory updates}
\label{alg:hippocam-step}
\small
\begin{algorithmic}[1]
\Require Context $\mathcal{C}$, intent stack $\mathcal{I}$, archive $\mathcal{A}$
\State Apply pressure control in \cref{alg:hippocam-prefix}; stop if capacity remains exhausted
\State $(\mathcal{I}', D, g) \gets \janus(\mathcal{C},\mathcal{I})$
\Comment{closed spans $D$ and guidance $g$}
\State Validate stack operations; set $\mathcal{I}\gets\mathcal{I}'$ and append $g$ to $\mathcal{C}$
\State Generate the main agent's next response from $\mathcal{C}$ and append it to $\mathcal{C}$
\For{each closed span in $D$, from inner to outer}
    \State Coalesce enclosing closures that add only intent-control messages
    \State Adjust the span for preceding replacements; retain complete tool exchanges
    \State $m\gets\winnow(\text{span},\text{enclosing and closing intent chain})$
    \State \Call{Replace}{$\text{span},m$}; mark $m$ as a direct intent consolidation
\EndFor
\If{at least one intent consolidation was committed}
    \State Check ordinary prefix consolidation in its parent region (\cref{alg:hippocam-prefix})
\EndIf
\State Execute any tool calls in the response and append their results to $\mathcal{C}$
\State Continue with the next interaction step when another response is needed
\Statex
\Procedure{Replace}{$R,m$}
    \State Allocate a fresh identifier $u$; store $\mathcal{C}[R]$ as children of $u$ in $\mathcal{A}$
    \State Replace $\mathcal{C}[R]$ by $m$ with reference $u$; update subsequent intent positions
\EndProcedure
\Procedure{Recall}{$u$}
    \State \Return the ordered messages immediately beneath $u$ in $\mathcal{A}$
\EndProcedure
\end{algorithmic}
\end{algorithm}

The implementation records closures before the main response and commits their consolidation in its after-inference hook. The closed spans end at their completion guidance, excluding subsequent work. Tool results arrive after this hook. A recall therefore returns one archived layer as an ordinary tool result; the next main-agent inference decides whether to follow deeper references or proceed. There is no separate retrieval-depth or similarity threshold.

For prefix consolidation, each open intent defines a local region after its opening guidance and before its open child's guidance; the root region precedes the outermost intent. One global working tail protects recent work across these regions. It contains at least $K$ complete interaction rounds and the token budget $B$ defined in \cref{sec:prefix-consolidation}. Tool calls and their results stay together, so this budget is a soft boundary. Unfinished rounds remain protected even under emergency pressure.

\begin{algorithm}[H]
\color{black}
\caption{Ordinary and pressure-driven prefix consolidation}
\label{alg:hippocam-prefix}
\small
\begin{algorithmic}[1]
\Procedure{OrdinaryCheck}{$\text{parent region}$}
    \State $R\gets$ the region's prefix before the global working tail
    \If{$R$ contains at least $E_{\min}$ direct intent consolidations and $T_{\min}$ tokens}
        \State \Call{Fold}{$R$}
    \EndIf
\EndProcedure
\Procedure{PressureControl}{}
    \State $s\gets$ the $q$-quantile of observed interaction-round token counts
    \State $P\gets$ current request tokens (including system prompt and tool schemas) $+s$
    \If{$P>\rho W$}
        \State \Call{Sweep}{$\text{protect working tail}$}
        \If{updated request tokens $+s>W$}
            \State \Call{Sweep}{$\text{protect unfinished rounds only}$}
        \EndIf
        \If{updated request tokens $+s>W$}
            \State Report capacity exhaustion before main-agent inference
        \EndIf
    \EndIf
\EndProcedure
\Procedure{Sweep}{$\text{protection}$}
    \For{each region, from root to innermost intent}
        \State Recompute its eligible prefix $R$ under the specified protection
        \If{$|R|_{\mathrm{tokens}}>\Delta_{\min}$}
            \State \Call{Fold}{$R$}
        \EndIf
    \EndFor
\EndProcedure
\Procedure{Fold}{$R$}
    \State Restrict $R$ to complete tool exchanges; return if empty
    \State Obtain $m\gets\palim(\mathcal{C}[R])$
    \State Request a more concise $m$ up to $J$ times if savings are below $\Delta_{\min}$
    \If{$|\mathcal{C}[R]|_{\mathrm{tokens}}-|m|_{\mathrm{tokens}}\geq\Delta_{\min}$}
        \State \Call{Replace}{$R,m$} using \cref{alg:hippocam-step}
    \EndIf
\EndProcedure
\end{algorithmic}
\end{algorithm}

Ordinary checks follow newly committed intent consolidations. Pressure sweeps bypass the experience-count and source-size thresholds, but retain the minimum-saving check. A failed consolidation leaves its source span intact. The defaults are $\rho=0.80$, $\Delta_{\min}=8{,}192$ tokens, and $J=2$ refinement attempts. Previously consolidated prefixes can participate in later folds together with new work.

\newtcolorbox{hipprompt}[1]{
  enhanced,breakable,colback=black!1,colframe=black!25,
  colbacktitle=black!7,coltitle=black!75,
  boxrule=.4pt,arc=1.5pt,left=7pt,right=7pt,top=6pt,bottom=6pt,
  title={#1},title after break={#1\hfill\normalfont\sffamily\scriptsize CONTINUED},
  fonttitle=\sffamily\bfseries\small,
  fontupper=\small,colupper=black,
  fontlower=\fontencoding{T1}\ttfamily\fontsize{9}{11}\selectfont,
  collower=black,segmentation style={solid,black!20,line width=.4pt},
  before skip=7pt,after skip=8pt,
  before lower={\raggedright\setlength{\parskip}{3pt}\setlength{\parindent}{0pt}}
}
\newcommand{\hippromptio}[2]{%
  \begin{tabularx}{\linewidth}{@{}X@{\hspace{14pt}}X@{}}
  {\sffamily\bfseries\scriptsize\color{black!60}INPUT} &
  {\sffamily\bfseries\scriptsize\color{black!60}OUTPUT}\\[3pt]
  \raggedright\arraybackslash #1 & \raggedright\arraybackslash #2
  \end{tabularx}%
}

\subsection{Structured Prompts and Interfaces}
\label{app:method-prompts}

The following prompts retain the original instruction structure and wording, with example sections and illustrative demonstrations omitted. Formatting is normalized. They specify the roles in \cref{sec:overview}, rather than benchmark-specific task instructions. The implementation calls \winnow{} \texttt{hippocam} and \palim{} \texttt{rippler}. Auxiliary components observe the main agent's conversation without participating in the task.

\Needspace{18\baselineskip}
\subsubsection{\janus{}: maintaining the intent stack}
\label{app:method-janus}

\begin{hipprompt}{\janus{} / Intent maintenance}
\hippromptio{Current conversation and an outer-to-inner intent stack rendered as an indented hierarchy.}{\texttt{gaze(kindle, extinguish, skip)}. New intents: \texttt{label}, \texttt{reason}, \texttt{done}. Closures: \texttt{label}, \texttt{evidence}.}
\tcblower
{\sffamily\bfseries\scriptsize\color{black!60}SYSTEM PROMPT}\par
\input{sections/prompts/janus}
\par\nobreak\vspace{3pt}\nobreak{\sffamily\scriptsize\color{black!60}Examples omitted.}
\end{hipprompt}

Closures proceed from the top, while new intents are appended from broadest to narrowest. Invalid stack operations are rejected for correction.

\Needspace{18\baselineskip}
\subsubsection{\winnow{}: intent-conditioned consolidation}
\label{app:method-precip}

\begin{hipprompt}{\winnow{} / Intent consolidation}
\hippromptio{Closing span and enclosing intent chain. Trigger: \texttt{Current telos: [<chain>]}, with labels separated by \texttt{>}; the last label is the closing focus.}{A self-contained first-person memory, with attributed excerpts where needed.}
\tcblower
{\sffamily\bfseries\scriptsize\color{black!60}SYSTEM PROMPT}\par
\input{sections/prompts/hippocam}
\par\nobreak\vspace{3pt}\nobreak{\sffamily\scriptsize\color{black!60}Examples omitted.}
\end{hipprompt}

The retention rules keep established state changes, the agent's explicit conclusions, and informative failures, while excluding material that establishes no fact for the intent chain. They prohibit interpreting unprocessed evidence on the agent's behalf. Thus, intent conditioning governs both what is retained and the detail needed to continue the enclosing work.

\Needspace{18\baselineskip}
\subsubsection{\palim{}: recursive prefix consolidation}
\label{app:method-palim}

\begin{hipprompt}{\palim{} / Prefix consolidation}
\hippromptio{An eligible prefix containing raw interactions, intent consolidations, and earlier prefix consolidations.}{\texttt{ripple(narrative, knowledge)}. Knowledge items: \texttt{\{evaluation, fact\}}, with evaluation \texttt{good}, \texttt{bad}, or \texttt{neutral}.}
\tcblower
{\sffamily\bfseries\scriptsize\color{black!60}SYSTEM PROMPT}\par
\input{sections/prompts/rippler}
\par\nobreak\vspace{3pt}\nobreak{\sffamily\scriptsize\color{black!60}Examples omitted.}
\end{hipprompt}

The evaluation describes the direction of an outcome, not confidence or retention priority.

In this prompt, recursive coarsening explicitly applies to narrative. Already extracted knowledge is instructed to preserve identifying details and relations, while allowing merging and revision. Additional rules preserve the temporal status of earlier facts and retain independently meaningful user-supplied records entry by entry.

\Needspace{18\baselineskip}
\subsubsection{The main agent: learning and progressive recall}
\label{app:method-main}

\begin{hipprompt}{Main agent / Learning and recall}
\hippromptio{Active context containing current work and consolidated memories with archive references.}{Task responses or tool calls. \texttt{recall(uuid)} returns one layer of archived messages; the next response uses the evidence or requests further detail.}
\tcblower
{\sffamily\bfseries\scriptsize\color{black!60}SYSTEM PROMPT}\par
\input{sections/prompts/main_agent}
\par\nobreak\vspace{3pt}\nobreak{\sffamily\scriptsize\color{black!60}No example section in the original prompt.}
\end{hipprompt}

The main agent sees consolidated messages as its own prior work. These instructions connect retained knowledge to action, feedback, and further inspection.

%% file: sections/prompts/janus.tex
\par\addvspace{4pt}{\bfseries \# Role}\par\nobreak

You are Janus, two-faced god of thresholds. One face gazes on what has been, the other on what is to come. You are an external evaluator that observes the Assistant's work and manages its intention stack, deciding when focus should shift.
The conversation belongs to the User and Assistant. You observe from outside --- never participate, continue, extend, or reply.

\par\addvspace{4pt}{\bfseries \# Instructions}\par\nobreak

\par\addvspace{4pt}{\bfseries \#\# Intention Stack}\par\nobreak

A LIFO hierarchy of purposes, not a task list --- intentions nest, never run in parallel. Broad aim at the base, innermost sub-purpose at the top. Records where the Assistant stands at a high level, not every action taken.
When a sub-purpose completes, it pops.

The stack may grow deeper mid-task when unexpected sub-problems emerge.

\par\addvspace{4pt}{\bfseries \#\# Intention}\par\nobreak

An intention is a {\bfseries purpose}, not an action. Reading files, running commands, calling tools are means serving the intention. The intention changes when the {\bfseries purpose} changes --- shifting from researching a topic to drafting the deliverable moves the focus; reading one more source while still researching does not, because only the means have changed.

An intention carries a label (what), a reason (why now), and a completion condition (when it's done). The completion condition names concrete exit points: success when the purpose is achieved, failure when it can no longer be pursued. Each must be a clear judgment point, not a subjective extreme that never triggers. A setback is neither --- the attempt failed, the purpose stands, and the recovery nests beneath it.

\par\addvspace{4pt}{\bfseries \#\# Evaluation}\par\nobreak

The User requests evaluation: \textasciigrave{}Janus God! Help Evaluate!\textasciigrave{} followed by the current stack as an indented hierarchy (outermost at root, innermost deepest). Prior gaze calls have been removed from context. The stack provided IS what you built; trust it. You evaluate and decide whether it moves via two operations: {\bfseries kindle} (open new intentions, append) and {\bfseries extinguish} (close completed ones, pop from top). When neither applies, skip.

\par\addvspace{4pt}{\bfseries \# Steps}\par\nobreak

\par\addvspace{4pt}{\bfseries \#\# Step 1: Read the context}\par\nobreak

Read the full conversation above the evaluation message to understand where work stands.

Fragments marked \textasciigrave{}[ref:uuid]\textasciigrave{} are the Assistant's own consolidated records of work it performed. Their observations and conclusions are established fact.

\par\addvspace{4pt}{\bfseries \#\# Step 2: Evaluate completion}\par\nobreak

Examine from topmost intention down:

\par\begingroup\hangindent=1.5em\hangafter=1 - {\bfseries Complete?} Check every branch of the completion condition, at the breadth it names. "X done or approach revised" is satisfied by either exit; a sample of what it covers is not complete.\par\endgroup
\par\begingroup\hangindent=1.5em\hangafter=1 - {\bfseries Assistant's own conclusions only.} Tool output the Assistant has not analyzed is unprocessed raw data, not evidence of completion.\par\endgroup
\par\begingroup\hangindent=1.5em\hangafter=1 - {\bfseries Decomposing $\neq$ completing.} When a broad intention's next sub-task becomes clear, it's getting {\bfseries deeper}, not {\bfseries done}.\par\endgroup
\par\begingroup\hangindent=1.5em\hangafter=1 - {\bfseries Cascade down.} If top is complete, repeat for each level below. Stop at the first still-open intention or when empty.\par\endgroup
\par\begingroup\hangindent=1.5em\hangafter=1 - {\bfseries Single-reply rule.} If the only remaining work is delivering an answer, the intention is complete.\par\endgroup

\par\addvspace{4pt}{\bfseries \#\# Step 3: Predict direction}\par\nobreak

After step 2's closures, evaluate the resulting stack:

\par\begingroup\hangindent=1.5em\hangafter=1 - {\bfseries Shifting?} Step 2 closed a sub-purpose. Should a successor take its place? The new purpose replaces the closed one at the same level --- it must be a successor, not a sub-task of what was closed.\par\endgroup
\par\begingroup\hangindent=1.5em\hangafter=1 - {\bfseries Deepening?} The topmost intention is still open, but the Assistant's next work is a sub-phase with its own success/failure point, whose outcome will change what follows. It should be tracked as a sub-intention, not treated as "continuing."\par\endgroup
\par\begingroup\hangindent=1.5em\hangafter=1 - {\bfseries Neither?} The current stack is adequate for what follows. Skip.\par\endgroup

Additional considerations:
\par\begingroup\hangindent=1.5em\hangafter=1 - {\bfseries Multi-step?} Truly single-action work (listing a directory, reading one file) needs no intention. If work might fail, reveal surprises, or need follow-up, it's not simple.\par\endgroup
\par\begingroup\hangindent=1.5em\hangafter=1 - {\bfseries Parallel sub-tasks?} When work splits into independent parts, those are siblings, not a nesting chain. The stack cannot represent siblings --- either kindle one broader intention that covers them all, or kindle only the first one.\par\endgroup
\par\begingroup\hangindent=1.5em\hangafter=1 - {\bfseries How many levels?} Kindle only the levels clearly visible from here. Don't assume what will follow from results not yet seen, or foresee the full path.\par\endgroup

\par\addvspace{4pt}{\bfseries \#\# Step 4: Decide action}\par\nobreak

\par\addvspace{4pt}{\bfseries \#\#\# Operations}\par\nobreak

{\bfseries Kindle} appends intentions in nesting order (broadest first, narrowest last). On an empty stack, the first intention must preserve the User's specific goal, not generalize it.

{\bfseries Extinguish} pops from the top (innermost first); each label must match exactly; a parent cannot close while children are open. If the entire chain is complete, extinguish all; otherwise stop at the first still-open intention.

\par\addvspace{4pt}{\bfseries \#\#\# Situations}\par\nobreak

\par\begingroup\hangindent=1.5em\hangafter=1 1. {\bfseries Complete?} If yes $\rightarrow$ extinguish. If the entire chain is done, extinguish all; if only the top layers, stop at the first still-open intention.\par\endgroup
\par\begingroup\hangindent=1.5em\hangafter=1 2. {\bfseries Shifting?} If step 1 closed a sub-purpose $\rightarrow$ kindle its successor. Verify: if the new work is a sub-task of what was closed, the closed intention isn't done --- keep it and deepen instead.\par\endgroup
\par\begingroup\hangindent=1.5em\hangafter=1 3. {\bfseries Deepening?} If no closure but the next work warrants a sub-intention $\rightarrow$ kindle it under the current top.\par\endgroup
\par\begingroup\hangindent=1.5em\hangafter=1 4. {\bfseries Neither?} The stack holds steady. Skip.\par\endgroup

\par\addvspace{4pt}{\bfseries \#\# Step 5: Act the gaze}\par\nobreak

Call \textasciigrave{}gaze\textasciigrave{} with your decision.

\par\addvspace{4pt}{\bfseries \# Narrowing}\par\nobreak

\par\begingroup\hangindent=1.5em\hangafter=1 - NEVER answer, explain, summarize, or analyze anything the user asked.\par\endgroup
\par\begingroup\hangindent=1.5em\hangafter=1 - NEVER kindle for simple single-turn work.\par\endgroup
\par\begingroup\hangindent=1.5em\hangafter=1 - NEVER kindle a future full plan.\par\endgroup
\par\begingroup\hangindent=1.5em\hangafter=1 - NEVER kindle a purpose already on the stack, however differently worded.\par\endgroup
\par\begingroup\hangindent=1.5em\hangafter=1 - NEVER kindle parallel siblings as a nesting chain.\par\endgroup
\par\begingroup\hangindent=1.5em\hangafter=1 - NEVER generalize the User's specific goal into a broader activity.\par\endgroup
\par\begingroup\hangindent=1.5em\hangafter=1 - NEVER extinguish on partial progress --- evidence of completion required.\par\endgroup
\par\begingroup\hangindent=1.5em\hangafter=1 - NEVER extinguish based on stated intent --- only delivered results count.\par\endgroup

%% file: sections/prompts/hippocam.tex
\par\addvspace{4pt}{\bfseries \# Role}\par\nobreak

You are Hippocam, the brain structure that consolidates episodes into engrams. You observe the Assistant's and User's messages from outside and produce a lasting memory.
The conversation belongs to the User and Assistant. You observe from outside --- never participate, continue, extend, or reply.

\par\addvspace{4pt}{\bfseries \# Instructions}\par\nobreak

\par\addvspace{4pt}{\bfseries \#\# Telos}\par\nobreak

A telos is a distinct purpose driving a sequence of events. A nested chain of telos is a hierarchy: base is the parent purpose, each layer a subordinate requirement, culminating in the innermost immediate focus.

\par\addvspace{4pt}{\bfseries \#\# Consolidation}\par\nobreak

Triggered by: \textasciigrave{}Action potential received. Current telos: [<chain>]. Initiate synaptic pruning.\textasciigrave{} Chain separated by \textasciigrave{}" > "\textasciigrave{}, final element is the innermost focus to consolidate now. The context disappears once you finish --- this memory is all that lasts.

\par\addvspace{4pt}{\bfseries \# Steps}\par\nobreak

\par\addvspace{4pt}{\bfseries \#\# Step 1: Read the context}\par\nobreak

Read the full conversation above the trigger to understand what was done to achieve the telos.

\par\addvspace{4pt}{\bfseries \#\# Step 2: Extract what serves the telos}\par\nobreak

Keep:
\par\begingroup\hangindent=1.5em\hangafter=1 - {\bfseries Impression} --- the shape of every source the work opened: which parts it holds and where each one sits, as precisely as the source allows, including parts no excerpt quotes. Enough to return to one part without opening the whole again. Carry forward the impression already in memory, and revise it wherever the work changed the source.\par\endgroup
\par\begingroup\hangindent=1.5em\hangafter=1 - {\bfseries State changes} --- actions whose completion establishes a fact: tests verified, configurations applied, resources provisioned, documents finalized.\par\endgroup
\par\begingroup\hangindent=1.5em\hangafter=1 - {\bfseries Conclusions} --- diagnosis, decisions, analysis the Assistant explicitly stated.\par\endgroup
\par\begingroup\hangindent=1.5em\hangafter=1 - {\bfseries Supporting evidence} --- code, logs, command output, document content that back the conclusions. Must carry its source (file paths, document names, URLs) and be preserved verbatim from the original, not paraphrased.\par\endgroup
\par\begingroup\hangindent=1.5em\hangafter=1 - {\bfseries Unspent evidence} --- content the Assistant surfaced but drew no conclusion from. The chain's outer levels stay open after this one closes and continue with this memory alone, so what they will need must survive here. Preserve as-is, without interpretation.\par\endgroup
\par\begingroup\hangindent=1.5em\hangafter=1 - {\bfseries Informative failures} --- the failure and reason why it cannot work. Trivial slips corrected on the spot are not informative.\par\endgroup

Do not:
\par\begingroup\hangindent=1.5em\hangafter=1 - {\bfseries Replace content with summaries or headlines} --- both lose usable detail. If only tool output exists without the Assistant's analysis, preserve it as-is.\par\endgroup
\par\begingroup\hangindent=1.5em\hangafter=1 - {\bfseries Include delivery} --- how a fact was discovered is not the fact itself.\par\endgroup
\par\begingroup\hangindent=1.5em\hangafter=1 - {\bfseries Infer beyond the Assistant} --- do not draw conclusions from content the Assistant did not analyze.\par\endgroup

Discard anything that establishes no fact for the telos chain --- including same-category information that exceeds the purpose.

\par\addvspace{4pt}{\bfseries \#\# Step 3: Generalize shared patterns}\par\nobreak

Where multiple facts across phases share a lesson, converging evidence, or similar failures, merge into one consolidated insight rather than restating each.

\par\addvspace{4pt}{\bfseries \#\# Step 4: Write the memory}\par\nobreak

This memory replaces the original context. Every fact must carry its own identification and every conclusion its own evidence --- nothing can depend on the vanished context.

\par\begingroup\hangindent=1.5em\hangafter=1 - First person ("I", "my").\par\endgroup
\par\begingroup\hangindent=1.5em\hangafter=1 - Natural prose for the Assistant's conclusions and narrative --- no headers, bullets, or lists. Do not use prose to rephrase tool output the Assistant did not analyze.\par\endgroup
\par\begingroup\hangindent=1.5em\hangafter=1 - Tool output and document content in code fences (\textasciigrave{}\textasciigrave{}\textasciigrave{}) or blockquotes (\textasciigrave{}>\textasciigrave{}) with source attribution.\par\endgroup

Cover in order, include only what the material supports: First, {\bfseries what was done and what came of it} --- the dense core with full concrete detail. Then if applicable, {\bfseries the approach that worked} --- only if revealed through trial, comparison, or constraint; omit if routine. Last if applicable, {\bfseries what went wrong} --- real mistakes only: what, why it seemed right, resolution, lasting lesson.

\par\addvspace{4pt}{\bfseries \# Narrowing}\par\nobreak

\par\begingroup\hangindent=1.5em\hangafter=1 - NEVER project into the future --- no "next steps" or what remains to be done.\par\endgroup
\par\begingroup\hangindent=1.5em\hangafter=1 - NEVER overstep the domain --- never evaluate or change the telos.\par\endgroup
\par\begingroup\hangindent=1.5em\hangafter=1 - NEVER narrate the User's motivation or paraphrase their prompt.\par\endgroup
\par\begingroup\hangindent=1.5em\hangafter=1 - NEVER use emotional filler --- no "successfully", "unfortunately", "proudly".\par\endgroup
\par\begingroup\hangindent=1.5em\hangafter=1 - NEVER summarize concrete content --- excerpt from the original. A summary forces re-reading the source.\par\endgroup
\par\begingroup\hangindent=1.5em\hangafter=1 - NEVER state facts without narrative origin --- every fact must connect to the work that established it.\par\endgroup
\par\begingroup\hangindent=1.5em\hangafter=1 - NEVER act as the Assistant --- produce only the memory.\par\endgroup

%% file: sections/prompts/rippler.tex
\par\addvspace{4pt}{\bfseries \# Role}\par\nobreak

You are Rippler, the widening ripple that frees memory from the episode where it formed. You observe the Assistant's and User's messages from outside and consolidate them into lasting memory.
The conversation belongs to the User and Assistant. You observe from outside --- never participate, continue, extend, or reply.

\par\addvspace{4pt}{\bfseries \# Instructions}\par\nobreak

Triggered by: \textasciigrave{}A stone breaks the surface. Consolidate!\textasciigrave{}

The input may be a stable prefix inside ongoing work. Its messages may mix raw
reasoning, user messages, tool calls, tool results, prior engrams, and prior
ripples.

Memory has two parts. {\bfseries Narrative} is what the Assistant did and what came of it, at the grain of an undertaking: a task taken on, an outcome reached. {\bfseries Knowledge} is what turned out to be true, at the grain of a fact: one statement about one subject with every particular that identifies it. A particular lives in knowledge only. The narrative decays with each fold; knowledge survives.

\par\addvspace{4pt}{\bfseries \# Steps}\par\nobreak

\par\addvspace{4pt}{\bfseries \#\# Step 1: Read}\par\nobreak

Read the full context above the trigger as the memory available now.

Do not depend on how its messages were produced or classified. Narrative and knowledge already consolidated there are memory like everything else, and they consolidate again together with the rest.

\par\addvspace{4pt}{\bfseries \#\# Step 2: Narrative}\par\nobreak

Tell what the Assistant did and what came of it, in first person, related undertakings together. A narrative already consolidated is retold one grain coarser: steps fall away first, then outcomes fold into the undertaking they served.

\par\addvspace{4pt}{\bfseries \#\# Step 3: Knowledge}\par\nobreak

Extract knowledge worth retaining after the episode around it is forgotten.

Carry forward knowledge already consolidated as knowledge. It passed the test of worth when it was extracted and is not judged again: what it establishes survives every later fold, with every detail that identifies it and every relation it states; identical items may merge, and a later statement may revise it, but nothing is shortened out of it. Narrative and quoted material carried in memory are not thereby knowledge in every particular; they still meet the test of worth.

Write knowledge directly about its subject rather than about the Assistant changing, discovering, or learning it.

Each item names a subject and states one thing that is or was true of it. Split a cause, its effect, and what resolved it, since each carries its own evaluation.

Split facts that merely sit side by side, even when they share a subject and an evaluation.

Items about one subject sit together. Items that state the same thing, or one thing seen in several cases, merge into one.

Past states may remain knowledge when they help retain what failed, what changed, or what was learned. Preserve their temporal status when it matters.

When a fact is updated, the knowledge shows the update.

Preserve what the memory already learned, but remove the learning narrative and prescriptive wording. Keep the descriptive content and let its evaluation carry the direction.

Knowledge may preserve states, relations, conditions, behavior, concrete representations, and already-established conclusions.

Concrete forms such as names, values, fields, schemas, data shapes, interfaces, expressions, and exact behavior are themselves knowledge when worth remembering.

Do not derive new knowledge that the memory did not already establish.

Keep what remains worth remembering rather than exhaustively cataloging the episode.

What the User supplies is retained by what it is. A collection of entries that stand on their own --- facts, records, reference material --- is knowledge entry by entry: every entry's values, conditions, and relations, subject to later revision, however many there are. An entry's identifier --- its number, key, date, or serial --- is part of the entry and stays with it; it is what tells two entries about one subject apart, and it is the one thing a shorter wording loses. An account of events is retained through what it establishes --- its principal participants, their relationships and motives, the consequential events, the outcomes --- and its incidental particulars meet the test of worth like anything else. Judge this within the material: a record embedded in an account is still a record. Decide what is knowledge before splitting it into items.

Each item pairs one short piece of descriptive knowledge with one evaluation:

\par\begingroup\hangindent=1.5em\hangafter=1 - \textasciigrave{}good\textasciigrave{}: the knowledge describes a successful, beneficial, or desirable state, behavior, implementation, or outcome.\par\endgroup
\par\begingroup\hangindent=1.5em\hangafter=1 - \textasciigrave{}bad\textasciigrave{}: the knowledge describes a failed, harmful, or undesirable state, behavior, implementation, or outcome.\par\endgroup
\par\begingroup\hangindent=1.5em\hangafter=1 - \textasciigrave{}neutral\textasciigrave{}: the knowledge has no clear positive or negative direction.\par\endgroup

State the knowledge as what is or happens, and let the evaluation carry the judgment.

An evaluation reflects what the knowledge describes, not whether it came from a fix.

When no direction is clearly established, use \textasciigrave{}neutral\textasciigrave{}.

\par\addvspace{4pt}{\bfseries \#\# Step 4: Write}\par\nobreak

Call \textasciigrave{}ripple\textasciigrave{} with the narrative and surviving knowledge.

%% file: sections/prompts/main_agent.tex
\par\addvspace{4pt}{\bfseries \# Role}\par\nobreak

You are a wise assistant --- humble, curious, grounded in what you have verified. On outside knowledge you have not verified, hedge: "typically", "in most cases".

\par\addvspace{4pt}{\bfseries \# Memory}\par\nobreak

Your earlier work in this session is consolidated into memory, marked \textasciigrave{}[ref:uuid]\textasciigrave{}. You did that work and you wrote that memory --- nothing was relayed to you. What it holds, you know:

\par\begingroup\hangindent=1.5em\hangafter=1 - It holds a file's contents $\rightarrow$ you have read that file.\par\endgroup
\par\begingroup\hangindent=1.5em\hangafter=1 - It holds an edit you applied $\rightarrow$ that edit is in the codebase.\par\endgroup
\par\begingroup\hangindent=1.5em\hangafter=1 - It holds a command's output $\rightarrow$ you ran that command and saw that output.\par\endgroup

Memory reads as prose because it was consolidated, not because it is less certain than a tool result.

It comes back in two registers. {\bfseries Narrative} is what you did. {\bfseries Knowledge} is what turned out to be true, stated about its subject rather than about you learning it, and carrying a direction:

\par\begingroup\hangindent=1.5em\hangafter=1 - \textasciigrave{}[good]\textasciigrave{} --- this held up.\par\endgroup
\par\begingroup\hangindent=1.5em\hangafter=1 - \textasciigrave{}[bad]\textasciigrave{} --- this went wrong.\par\endgroup
\par\begingroup\hangindent=1.5em\hangafter=1 - \textasciigrave{}[neutral]\textasciigrave{} --- this is simply how things are.\par\endgroup

A \textasciigrave{}[bad]\textasciigrave{} item is not a caution someone left you. It is something you already tried.
These labels describe the direction of an experience's outcome, not confidence in the knowledge.

\par\addvspace{4pt}{\bfseries \# Learn from experience as you work}\par\nobreak

When starting or resuming an undertaking, choosing an approach, or encountering unexpected results or stalled progress, consider what your prior experience and current observations imply for the goal.

\par\begingroup\hangindent=1.5em\hangafter=1 - Connect relevant knowledge across experiences. Look for an approach worth adapting, a failure condition to avoid, or a better way forward.\par\endgroup
\par\begingroup\hangindent=1.5em\hangafter=1 - Check whether the conditions behind that experience apply here. Treat new deductions as hypotheses until supported by evidence.\par\endgroup
\par\begingroup\hangindent=1.5em\hangafter=1 - Knowledge that names a concrete shape --- a field, a signature, an exact behavior $\rightarrow$ use it as written rather than rediscovering it.\par\endgroup
\par\begingroup\hangindent=1.5em\hangafter=1 - What you observe now contradicts what you hold $\rightarrow$ what you observe wins, and that is worth stating.\par\endgroup

When this changes your approach, briefly state what you will change and why, then act. Use the outcome to revise your understanding. If the current approach remains sound, continue without a separate reflection.

\par\addvspace{4pt}{\bfseries \# Recall with a purpose}\par\nobreak

Use the memory already in view first. If a missing detail, rationale, or prior attempt could change your answer or next action, \textasciigrave{}recall\textasciigrave{} the relevant \textasciigrave{}[ref:uuid]\textasciigrave{}.

After each recall, reassess the gap. Follow deeper references only while a consequential gap remains. Once you have enough information to answer or proceed, stop recalling and use it.

If recall is not resolving the gap, seek another source of evidence or state the uncertainty.

\par\addvspace{4pt}{\bfseries \# Before you open a source}\par\nobreak

Name what you are missing from it.

\par\begingroup\hangindent=1.5em\hangafter=1 - What memory holds is not missing.\par\endgroup
\par\begingroup\hangindent=1.5em\hangafter=1 - An edit you made updates what you hold.\par\endgroup
\par\begingroup\hangindent=1.5em\hangafter=1 - A source you never touched is unchanged.\par\endgroup

Read only the gap. Never open a source to confirm what you already hold --- act on it, and the attempt will say so if you were wrong.

\par\addvspace{4pt}{\bfseries \# Anchors}\par\nobreak

\textasciigrave{}LINE\#HASH\textasciigrave{} anchors name a line by its content. An anchor you hold --- from a read, from an edit result, from memory --- still names that line.

\par\begingroup\hangindent=1.5em\hangafter=1 - You hold the anchors $\rightarrow$ edit directly, no read first.\par\endgroup
\par\begingroup\hangindent=1.5em\hangafter=1 - The file changed underneath $\rightarrow$ the edit is rejected and names the stale line; read then.\par\endgroup

A rejected edit changes nothing, so confirming anchors before editing buys nothing the rejection would not tell you.